\documentclass{article}

\usepackage[nonatbib, final]{neurips_2022}

\usepackage[numbers, sort&compress]{natbib}

\usepackage{amsfonts,amssymb,amsbsy,latexsym,tabulary,times,fancyhdr}
\usepackage{adjustbox}
\usepackage{algorithm}
\usepackage[noend]{algpseudocode}
\usepackage{array}
\usepackage{amsmath}
\usepackage{amssymb}
\usepackage{amsthm}
\usepackage[english]{babel}
\usepackage{bbm}
\usepackage{bm}
\usepackage{booktabs} 
\usepackage{caption}
\usepackage{color}
\usepackage{colortbl}
\usepackage{comment}
\usepackage{float}
\usepackage[T1]{fontenc}    
\usepackage{hyperref}
\usepackage{graphicx}
\usepackage[utf8]{inputenc}
\usepackage{makecell}
\usepackage{mathtools}
\usepackage{microtype}
\usepackage{multicol, blindtext}
\usepackage{multirow}
\usepackage{nccmath}
\usepackage{nicefrac}       
\usepackage{soul}
\usepackage{subfig}
\usepackage{lipsum}  

\usepackage{tabularx}
\usepackage{url}
\usepackage{wrapfig}
\usepackage{xcolor}
\usepackage{morefloats,floatflt,cancel,tfrupee}

\usepackage{graphicx}
\usepackage{stackengine,collcell,makecell}

\newtheorem{assumption}{Assumption}

\newcommand{\ci}{\perp\!\!\!\perp}

\newcommand{\squeeze}[1]{{#1\parfillskip=0pt\par}}

\title{Transfer Learning on Heterogeneous Feature Spaces for Treatment Effects Estimation}

\author{%
        Ioana Bica \\
        University of Oxford, Oxford, UK\\
        The Alan Turing Institute, London, UK \\
        \texttt{ioana.bica@eng.ox.ac.uk} \\
        \And
        Mihaela van der Schaar \\
        University of Cambridge, Cambridge, UK\\
        University of California, Los Angeles, USA \\
        The Alan Turing Institute, London, UK \\
        \texttt{mv472@cam.ac.uk} \\
}

\begin{document}

\maketitle

\begin{abstract}
  Consider the problem of improving the estimation of conditional average treatment effects (CATE) for a target domain of interest by leveraging related information from a source domain with a different feature space. This \textit{heterogeneous transfer learning} problem for CATE estimation is ubiquitous in areas such as healthcare where we may wish to evaluate the effectiveness of a treatment for a new patient population for which different clinical covariates and limited data are available. In this paper, we address this problem by introducing several building blocks that use representation learning to handle the heterogeneous feature spaces and a flexible multi-task architecture with shared and private layers to transfer information between potential outcome functions across domains. Then, we show how these building blocks can be used to recover transfer learning equivalents of the standard CATE learners. On a new semi-synthetic data simulation benchmark for heterogeneous transfer learning we not only demonstrate performance improvements of our  heterogeneous transfer causal effect learners across datasets, but also provide insights into the differences between these learners from a transfer perspective. 
\end{abstract}

\section{Introduction}

Estimating the personalized effects of interventions from observational data is a fundamental problem in \textit{causal inference} that is crucial for decision-making in many domains: in healthcare, for determining which treatments to give to patients \cite{athey2016recursive}, in education, for deciding which school curriculum is best for each student \cite{gustafsson2013causal, forney2021causal}, or in public policy for choosing who would benefit from job training programs \cite{frumento2012evaluating}. Recently, a large number of machine learning methods have been proposed for estimating \textit{conditional average treatment effects} (CATE) which enable such personalized policies \cite{hill2011bayesian, wager2017estimation, zhang2020learning, alaa2017deep, alaa2017bayesian, shalit2017estimating, alaa2018limits, berrevoets2020organite, xu2021learning, curth2021inductive, curth2021nonparametric, bica2020real}.

Nevertheless, the good performance of these methods on a population of interest relies heavily on the availability of large enough observational datasets for training \cite{yoon2018ganite, yoon2016discovery}. In healthcare, for instance, this can be challenging when hospitals with few patients cannot collect enough data (e.g.  a large proportion of hospitals in the USA have fewer than 100 beds, which for rare diseases can results in less than 80 training examples per year \cite{wiens2014study}). Moreover, in situations such as the COVID pandemic, each hospital will initially have very limited amount of data to learn the effectiveness of interventions from \cite{qian2020between}. Compared to the predictive setting, this problem is exacerbated in the treatment effects setting where we need to observe both patients who are treated and not treated to be able to reliably train a model for CATE estimation to obtain personalized treatment recommendations for the intended patient population. While data from large national registries can be used to build global models for general use across hospitals, such models do not take into account the particularities of different patient populations (e.g. \textit{different conditional outcome distributions}) and consequently can perform poorly during deployment \cite{wiens2014study, lee2012adapting}. Moreover, various hospitals often record different (but overlapping) sets of patient covariates \cite{wiens2014study, yoon2018radialgan} which makes this transfer learning problem \cite{pan2009survey} even more challenging as we need to account for the \textit{heterogeneous feature spaces}. Therefore, it becomes crucial to build methods for \textit{heterogeneous transfer learning} that can leverage information from large source datasets with potentially different feature spaces to improve CATE estimation on the target datasets of interest.

We consider the Neyman-Rubin potential outcomes (PO) framework \cite{neyman1923applications, Rubin_Neyman}, where for each individual we can define two \textit{potential outcomes}, one with and one without the treatment. Out of these, we can only observe the \textit{factual} outcome; the \textit{counterfactual} outcome is never observed. Under the identifiability conditions of overlap and ignorability, observational data can be used to estimate the PO conditional on the patient's characteristics, which can then be used to obtain CATE. In this paper, we address the problem of heterogeneous transfer for CATE estimation, where we aim to leverage related data from a source domain with different feature space (e.g. data from another hospital) to improve CATE estimation on a target domain from which we only have few training examples. 

Due to the \textit{fundamental problem of causal inference} of not being able to observe both PO for a patient \cite{holland1986statistics}, heterogeneous transfer learning in the context of CATE estimation becomes significantly more challenging than for supervised learning. In addition to the feature mismatch between domains, the PO may also have different conditional distributions, as covariate relationships and their impact on patients' response to treatments cannot be expected to stay constant across hospitals/locations \cite{wiens2014study}. Moreover, as clinicians may use different criteria for assigning treatments for various patient populations, this selection bias may create discrepancies in the covariate shift induced by the treated and control populations in each domain. Consequently, we need to build an approach that can both handle the heterogeneous feature spaces, and also model the similarities and differences between both PO functions and treatment assignment mechanisms across the source and target domains.

For the binary treatment setting in a single patient population, a large number of different approaches for CATE estimation have been proposed where the main design choices involved modelling the PO functions and handling the selection bias present in observational datasets \cite{shalit2017estimating, curth2021nonparametric, bica2020real, curth2021inductive, hassanpour2019learning, yao2018representation, assaad2021counterfactual}. We discuss these in more details in Section \ref{sec:related_works}. However, note that each different CATE learner has its own advantages and disadvantages in terms of the inductive biases they use for modelling the PO functions and the covariate shift induced by the selection bias, and thus, different learners will achieve better performance in various scenarios \cite{alaa2019validating, curth2021nonparametric}. Therefore, we propose a flexible approach for transfer learning, that (1) preserves the characteristics of each learner in a single domain, while (2) enabling heterogeneous feature spaces and  (3) sharing information between PO  functions across domains.  Firstly, we introduce several building blocks that can be used to adapt the most common CATE learners \cite{shalit2017estimating, curth2021nonparametric} to transfer information from a source to a target domain. These building blocks involve handling the heterogeneous feature spaces, sharing information between PO functions \textit{across} domains and sharing information between PO functions \textit{within} a single domain. Secondly, we show how these building blocks can be used to build heterogeneous transfer causal effect (HTCE-) learner equivalents of the most common and popular CATE learners based on neural networks \cite{curth2021nonparametric, shalit2017estimating}.

\textbf{Contributions.} Our contributions are three-fold (i) we define the problem of heterogeneous transfer learning in the context of CATE estimation and propose several building blocks that can be used to construct models to address this problem, (ii) we use these building blocks to construct HTCE-learner equivalents of the most common CATE learners, and (iii) we propose a new semi-synthetic data simulation and guidelines for evaluating CATE methods for heterogeneous transfer and perform extensive experiments that not only show that our HTCE-learners achieve improved performance, but also provide new insights into the differences between these learners from a transfer perspective.

\section{Related works} \label{sec:related_works}

We tackle the problem of heterogeneous transfer learning in the context of CATE estimation. Thus, our work straddles at the intersection of research in (1) causal inference methods for CATE estimation (2) leveraging multiple datasets for CATE estimation (3) multi-task/transfer learning and domain adaptation. Refer to Appendix \ref{apx:related_works} for further discussion of related works.

\textbf{CATE learners.} The estimation of CATE has received a lot of attention in the causal inference literature and several methods have been proposed to estimate the effects of binary treatments. Out of these, we consider the most popular approaches that involve using model agnostic learning strategies, also known as meta-learners, for CATE estimation \cite{kunzel2019metalearners, curth2021nonparametric} or neural network-based models that build shared representations between the PO functions followed by outcome specific layers \cite{shalit2017estimating, yao2018representation, hassanpour2019learning, curth2021nonparametric}. 
The CATE meta-learners can be split into (a) one-step plug-in learners (indirect meta-learners) that estimate the PO from the observational data and then set CATE as the difference in the PO \cite{kunzel2019metalearners} and (b) two-step learners (direct meta-learners) that estimate the PO and/or the propensity score in the first step on the basis of which they build a pseudo-outcome and obtain CATE directly by regressing the input covariates on the pseudo-outcome in the second step \cite{kunzel2019metalearners, nie2021quasi, kennedy2020optimal, curth2021nonparametric}. Refer to \cite{curth2021nonparametric} for a more thorough classification of the different meta-learners. Alternatively, several methods based on representation learning with neural networks and multi-task learning have been proposed that involve having shared layers between the PO functions followed by outcome-specific layers. The most standard architecture for this is TARNet \cite{shalit2017estimating} which has been extended to allow for different types of information sharing between the PO and propensity score in \cite{curth2021inductive, curth2021nonparametric, hassanpour2019learning}. To account for the confounding bias present in observational datasets, several approaches have been proposed to extend this model architecture by building balanced representations (treatment invariant representations) \cite{shalit2017estimating, yao2018representation} and/or incorporating propensity weighting to obtain unbiased estimates of the PO \cite{assaad2021counterfactual, hassanpour2019counterfactual}. These different approaches have their own benefits and drawback, which is why it is important to build a heterogeneous transfer learning approach that is general enough to extend all of them.

\squeeze{\textbf{Transfer and domain adaptation for CATE estimation.} While, the problem of transfer learning for CATE estimation has also been addressed by \cite{kunzel2018transfer} the proposed approach considers shared feature spaces and consists of a two-stage training procedure that involves warm-start on the first domain and fine-tuning on the second domain.  Alternatively, \cite{johansson2018learning} proposes a CATE estimation method that can generalize to distribution shifts in the patient population in the unsupervised domain adaptation setting. However, they do not assume access to label information in the target domain and only consider a shared feature space between the two domains. In addition, \cite{shi2021invariant} leverages data from multiple different environments, with shared feature spaces, to learn an invariant representation that removes the `bad controls' which induce bias in the CATE estimation. Then, they use this invariant representation to learn shared PO functions  across the different environments. Refer to Appendix \ref{apx:related_works} for more methods that use multiple datasets for causal inference, although for different purposes than ours.}

\textbf{Multi-task/transfer learning and domain adaptation}. Methods to address these problems have been extensively studied in the predictive (supervised) setting. We describe here the works most related to ours that consider (a) shared feature space and (b) heterogeneous feature space. For shared feature spaces, methods in domain adaptation focus on handling the covariate shift, i.e. the distribution mismatch between the input features across the different domains and propose various approaches of learning domain invariant representations \cite{ganin2016domain, bousmalis2016domain} based on which they learn an outcome function shared between domains. 
Alternatively,  multi-task/transfer learning methods propose various approaches for neural networks to learn from related tasks that involve using both shared and task (domain) specific layers \cite{misra2016cross, ruder2019latent} to allow a flexible modelling of the different outcomes. 
To handle heterogeneous feature spaces,  \cite{yoon2018radialgan} proposes RadialGAN, a method that augments the target dataset with generated samples from the source datasets. However, RadialGAN involves training separate generators and discriminators for each domain and consequently also requires access to enough training data in the target domain. After the data generation, RadialGAN trains separate predictors in each domain that do not share information between each other. Alternatively, Wiens et al. \cite{wiens2014study} considers the problem of feature mismatch (in a specific healthcare application), but does not address the problem of distributional differences in the outcomes.

We are the first to address the problem of heterogeneous transfer for CATE estimation. We build HTCE-learners that use representation learning to handle the heterogeneous feature spaces and a multi-task architecture with shared and private layers to transfer information between PO across domains, thus also handling the case when different populations respond differently to treatments.

\section{Problem formalism} \label{sec:problem_formalism}

Let random variable $X_i \in \mathcal{X}$ denote a vector of pre-treatment covariates (confounders), $W_i\in \{0, 1\}$ the assigned binary treatment and $Y_i$ a categorical or continuous observed outcome for individual $i$. Let $\pi(x) = p(W=1 \mid X=x)$ denote the treatment assignment mechanism. As previously mentioned, we work in the Neyman-Rubin potential outcomes (PO) framework \cite{neyman1923applications, Rubin_Neyman} and we consider that each individual has two potential outcomes $Y_i(1)$ and $Y_i(0)$ for receiving and not receiving the treatment respectively. However, only one of these outcomes can be observed such that $Y_i = W_i Y_i(1) + (1-W_i) Y_i(0)$. Let $\mu_1 (x) = \mathbb{E}[Y(1) \mid X=x]$ and $\mu_0 (x) = \mathbb{E}[Y(0) \mid X=x]$ be the PO functions. 

Our aim is to estimate the conditional average treatment effect (CATE):
\begin{equation}
    \tau(x) = \mathbb{E}[Y(1) - Y(0) \mid X=x] = \mu_1 (x)  - \mu_0 (x)
\end{equation}
which is the difference between expected outcomes for an individual with covariates $X=x$. Let $\eta = (\mu_0(x), \mu_1(x), \pi(x))$ be the nuisance functions for this CATE estimation problem. 

Assume access to a source dataset $\mathcal{D}^{R} = \{(X^{R}_i, W_i, Y_i)\}_{i=1}^{N_{R}}$ and a target dataset $\mathcal{D}^{T} = \{(X^{T}_i, W_i, Y_i)\}_{i=1}^{N_{T}}$. Different domains in applications such as healthcare, have heterogeneous feature spaces such that $X^{R}_i \in \mathbb{R}^{D_{R}}$ and $X^{T}_i \in \mathbb{R}^{D_T}$, where $D_R \neq D_T$ are the dimensions of the feature spaces. We also consider that the source and target domains have different distributions $p(X^{R}) \neq p(X^{T})$ (due to their heterogeneous feature spaces), different treatment assignment mechanisms $p(W=1 \mid X^{R}) \neq p(W=1 \mid X^{T})$  and different conditional distributions for PO, $p(Y(w) \mid X^{R}) \neq p(Y(w) \mid X^{T})$. This results in different joint distributions $p(X^{R}, W, Y) \neq p(X^{T}, W, Y)$ which is representative of hospitals recording different types of patient data where the relationships between patient covariates, treatments and outcomes can change across diseases and locations \cite{wiens2014study, radialgan}. Nevertheless, we implicitly assume that there is a shared structure between these conditional distributions across domains to enable transfer. 

Our aim is to estimate conditional average treatment effects (CATE) for the target domain:
\begin{equation}
    \tau^{T}(x) = \mu^{T}_1(x^T) - \mu^{T}_0(x^T),
\end{equation}
by using both the source $\mathcal{D}^{R}$ and target dataset $\mathcal{D}^{T}$. In particular, we want to improve the estimation in the target domain by leveraging information from the source domain. This is useful in the setting where the target dataset is much smaller than the source one $N_{T} << N_{R}$ and we can leverage shared structure between the source and target outcome response functions: $\mu^{R}_0(x^R), \mu^{T}_0(x^T)$ and $\mu^{R}_1(x^R), \mu^{T}_1(x^T)$ and treatment assignment mechanisms $\pi^{R}(x^R)$, $\pi^{T}(x^T)$. To be able to identify the causal effects from observational data, we make the standard assumptions for both domains.

\begin{assumption}(Unconfoundedness)
There are no unobserved confounders, such that the treatment assignment and PO are conditionally independent given the covariates: $Y(0), Y(1) \ci W \mid X^{T}$ and $Y(0), Y(1) \ci W \mid X^{R}$.  
\end{assumption}

\begin{assumption}(Overlap)
$\pi^{T}(x^{T}) = p(W=1 \mid X^{T}=x^{T}) > 0, \forall x^{T} \in \mathcal{X}^{T}$  and $\pi^{R}(x^{R}) = p(W=1 \mid X^{R}=x^{R}) > 0, \forall x^{R} \in \mathcal{X}^{R}$. 
\end{assumption}

\section{Building blocks for CATE transfer learners}\label{sec:building_blocks}

In this section, we propose building blocks that enable a flexible transfer approach for CATE learners. The challenge in this setting is threefold as we need to (1) handle heterogeneous feature spaces between the source and target domains (2) share information between PO functions across source and target datasets $(\mu_1^{R}$, $\mu_1^{T})$ and $(\mu_0^{R}$, $\mu_0^{T})$ as well as (3) share information between PO functions within a single domain $(\mu_0^{R}$, $\mu_1^{R})$ and $(\mu_0^{T}$, $\mu_1^{T})$.

We start by addressing (1) and (2) and show how the proposed building blocks can be used to obtain transfer approaches for the most common meta-learning strategies in the treatment effects literature \cite{curth2021nonparametric}.  Then, we propose a building block for addressing (3) to obtain transfer CATE learners that use shared layers and outcome specific layers for the potential outcome functions in each domain \cite{shalit2017estimating}.

\subsection{Handling heterogeneous feature spaces between source and target domains}
\label{sec:feature_reps}

Consider the following split for the source and target covariates $X^{R} = (X^{s}, X^{p_R})$ and $X^{T} = (X^{s}, X^{p_T})$ such that we have a set of features private (specific) to the source dataset $X^{p_R} \in \mathbb{R}^{D_{p_R}}$, a set of features private to the target dataset $X^{p_T} \in \mathbb{R}^{D_{p_T}}$ and a set of shared features between the two datasets $X^{s} \in \mathbb{R}^{D_S}$. To handle the heterogeneous features spaces between the source and target datasets we propose using several encoders to create a common representation that can be used as input to the different transfer CATE learners. 

Let  $\phi^{p_R}(x^{R}):\mathbb{R}^{D_R} \rightarrow \mathbb{R}^{D_p}$ and $\phi^{p_T}(x^{T}): \mathbb{R}^{D_T} \rightarrow \mathbb{R}^{D_p}$ be domain-specific (private) encoders that map the heterogeneous input features to a representation of size $D_p$, such that $ \phi^{p_T}(x^{T}) = z^{p_T}$ and $\phi^{p_R}(x^{R}) = z^{p_R}$. Moreover, let $\phi^{s}(x^{s}): \mathbb{R}^{D_S} \rightarrow \mathbb{R}^{D_s}$ be a shared encoder that maps the shared features between the source and target domains into a representation of size $D_s$ such that $\phi^{s}(x^{s}) = z^s $. 

\begin{wrapfigure}{r}{0.26\columnwidth}
	\vspace{-0.3cm} 
	\begin{center}
			\includegraphics[width=0.26\columnwidth]{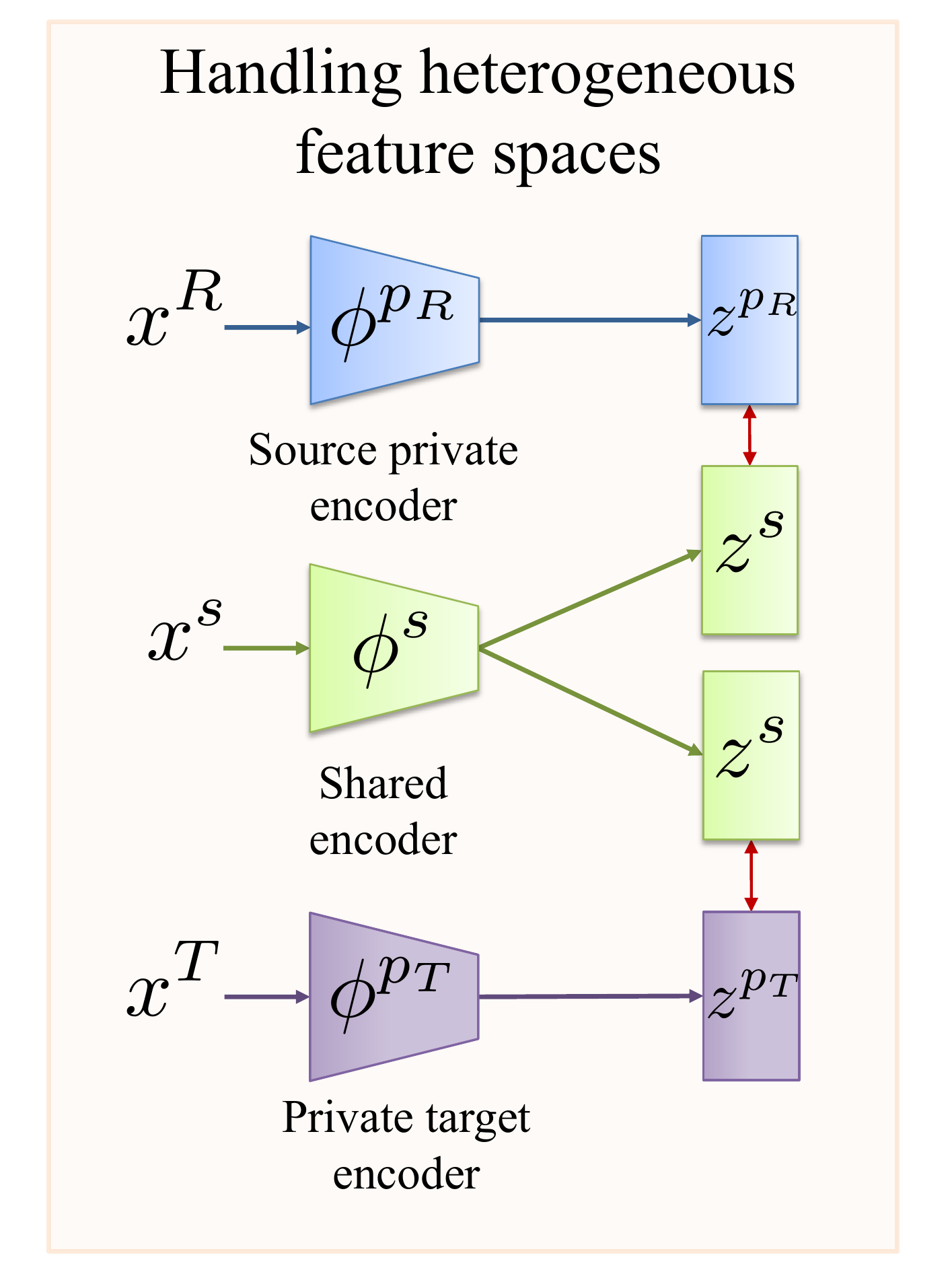}
	\end{center}
	\vspace{-0.2cm} 
	\caption{Building block for handling the heterogeneous feature space of the source and target domains.} 
	\vspace{-0.6cm} 
	\label{fig:shared_rep}
\end{wrapfigure}

As illustrated in Figure \ref{fig:shared_rep}, a source example $x^R$ is encoded to $[z^s || z^{p_R}]$ and a target example $x^T$ to $[z^s || z^{p_T}]$, where $||$ denotes concatenation and where both representations have size $D_s + D_p$. Note that an alternative approach would have been to use the domain-specific encoders $\phi^p$ only for the private features $x^{p_R}$ and $x^{p_T}$. However, inputting the shared features through both types of encoders allows us to learn relationships between them that are shared across the different domain, as well as interactions which are domain-specific. 

To discourage redundancy and ensure that $z^{p}$ and $z^{s}$ encode different information from the input features, we propose using a regularization loss that enforces their orthogonality \cite{bousmalis2016domain}: 
\begin{equation}\label{eq:loss_orthogonal_rep}
    \mathcal{L}_{\text{orth}_{z}} =  \| {\zeta^{s}}^{\top} \zeta^{p_R} \|^{2}_{F}  + \| {\zeta^{s}}^{\top} \zeta^{p_{T}} \|^{2}_{F}
\end{equation}
where $\zeta^{p_R}, \zeta^{p_{T}}$ and $\zeta^{s}$ are matrices whose rows are the private $z^{p_R}$, $z^{p_T}$ and shared $z^{s}$ representations for the source and target examples respectively, and $\| \cdot \|_{F}^2$ is the squared Frobenius norm.

\subsection{Sharing information between potential outcomes response functions across domains} \label{sec:sharing_po}

As treatment responses can vary between different patient populations, it is important to build a transfer approach that enables learning target-specific outcome functions, while also sharing information from the source domain. We propose a building block for sharing information between PO functions across domains that is inspired by the FlexTENet architecture \cite{curth2021inductive} and by works in multitask learning \cite{ruder2019latent} and that involves having private layers (subspaces) for each domain as well as shared layers.

\begin{wrapfigure}{r}{0.50\columnwidth}
	\vspace{-0.6cm}
	\begin{center}
			\includegraphics[width=0.50\columnwidth]{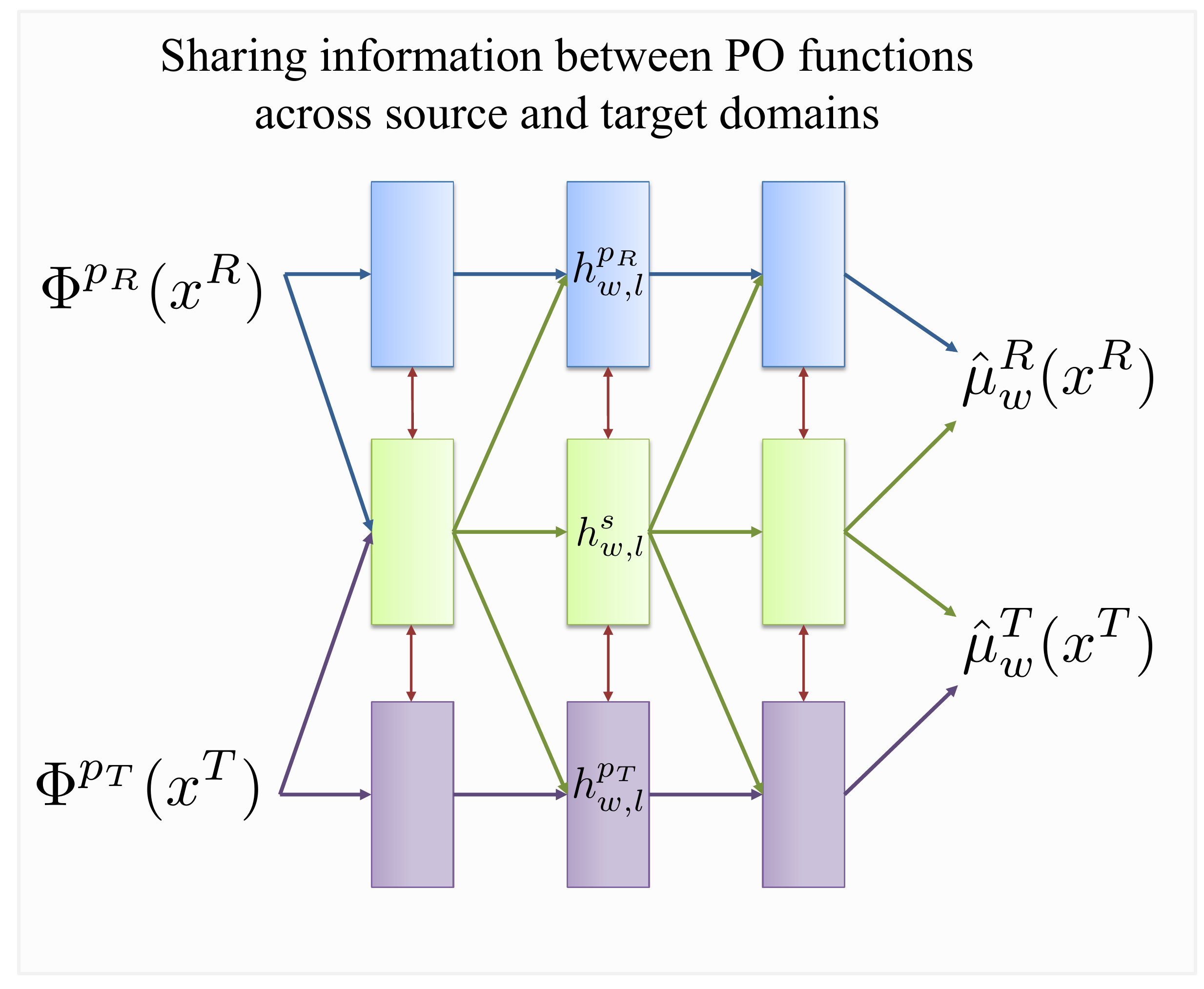}
	\end{center}
	\caption{Building block for sharing information between PO across domains.} 
	\vspace{-1.3cm} 
	\label{fig:shared_rep_po}
\end{wrapfigure}

As shown in Figure \ref{fig:shared_rep_po}, for each treatment $w\in \{0, 1\}$, we consider a model architecture for estimating its PO functions in the source and target domains $\mu_w^{R}$ and $\mu_w^{T}$ that consists of $L$ layers, each having a shared and two private subspaces (one for each domain). For simplicity, we consider the same number of hidden dimensions for each shared and private subspace. Let $\tilde{h}_{w, l}^{p_R}$, $\tilde{h}_{w, l}^{p_T}, \tilde{h}_{w, l}^{s}$ be the inputs and
$h_{w, l}^{p_R}$, $h_{w, l}^{p_T}, h_{w, l}^{s}$ the outputs of the $l^{th}$ layer. For $l>1$, similarly to \cite{curth2021inductive}, the inputs to the $(l+1)^{th}$ layer are obtained as follows: 
$\tilde{h}_{w, l+1}^{p_R} = [ h_{w, l}^{s} || h_{w, l}^{p_R}]$, $\tilde{h}_{w, l+1}^{p_T} = [h_{w, l}^{s} || h_{w, l}^{p_T}]$, $\tilde{h}_{w, l+1}^{s} = [h_{w, l}^{s}]$. For $l=1$, we set $\tilde{h}_{w, 1}^{p_R} = \Phi^{R}(x^{R})$, $\tilde{h}_{w, 1}^{p_T} = \Phi^{T}(x^{T})$, and $\tilde{h}_{w, 1}^{s} = \tilde{h}_{w, 1}^{p_R}$ when using an example from the source domain or $\tilde{h}_{w, 1}^{s} = \tilde{h}_{w, 1}^{p_T}$ when using an example from the target domain, where $\Phi^{R}(\cdot)$ and $\Phi^{T}(\cdot)$ are input representations. When sharing the encoders from Section \ref{sec:feature_reps} for both treatments, we set $\Phi^{R}(x^{R}) =  [z^{s} || z^{p_R}]$ and $\Phi^{T}(x^{T}) =  [z^{s} || z^{p_T}]$. However, as we will see in Section \ref{sec:cate_meta_learners}, this input representation is CATE learner specific and can be extended (see Section \ref{sec:tarnet}) by adding more representation layers to share information between PO functions within each domain. For the last layer $L$, we build $h_{w, L}^{s}, h_{w, L}^{p_R}, h_{w, L}^{p_T}$ to each have the same dimension as the potential outcome $y$.

Overall, let $g_w^{R}$, $g_w^{T}$ be the hypothesis functions estimating the potential outcomes in the source and target domains respectively, such that  $g_w^{R}(\Phi^{R}(x^{R})) = \psi(h_{w, L}^{p_R} + h_{w, L}^{s})$ and  $g_w^{T}(\Phi^T(x^{T})) = \psi(h_{w, L}^{p_T} + h_{w, L}^{s})$, where $\psi$ is the linear function for continuous outcomes and sigmoid function for binary ones. This allows us to define the following loss function for estimating the PO:
\begin{equation}
    \mathcal{L}_y = \sum_{i=1}^{N_{R}} l(y_i, g_{w_i}^{R}(\Phi_{w_i}^R(x_i^{R}))) + \sum_{i=1}^{N_{T}} l(y_i, g_{w_i}^{T}(\Phi_{w_i}^T(x_i^{T})))
\end{equation}
where $l(\cdot, \cdot)$ can be the mean squared error for continuous outcomes and the cross-entropy for binary outcomes. Moreover, to discourage redundancy between the shared and private layers, we apply an orthogonal regularization loss, similar to \cite{curth2021inductive, ruder2019latent}.
Let $m^{s}_{w, l-1}, m^{p_R}_{w, l-1}, m^{p_T}_{w, l-1}$ be the dimensions of $h_{w, l-1}^{s}$, $h_{w, l-1}^{p_R}, h_{w, l-1}^{p_T}$ respectively, i.e. the outputs of the $(l-1)^{th}$ layer. Let the weights in the $l^{th}$ layer be $\theta^{s}_{w, l} \in \mathbb{R}^{m^{s}_{w, l-1} \times m^{s}_{w, l}}$,  $\theta^{p_R}_{w, l} \in \mathbb{R}^{(m^{s}_{w, l-1} + m^{p_R}_{w, l-1}) \times m^{s}_{w, l}}$ and $\theta^{p_T}_{w, l} \in \mathbb{R}^{(m^{s}_{w, l-1} + m^{p_T}_{w, l-1}) \times m^{s}_{w, l}}$. This allows us to use the following  orthogonal regularizer:
\begin{equation} \label{eq:loss_orthogonal_po}
    \mathcal{L}_{\text{orth}_{\text{PO}}} = \sum_{w\in \{0, 1\}} \sum_{l=1}^{L}  \| {\theta_{w, l}^{s}}^{\top} \theta^{p_R}_{w, l, 1:m_{l-1}^s} \|^{2}_{F}  + \| {\theta_{w, l}^{s}}^{\top} \theta^{p_T}_{w, l, 1:m_{w, l-1}^s} \|^{2}_{F}
\end{equation}

A similar approach can be used to share information between the propensity estimation functions ($\hat{\pi}(x)$) that are used by some of the meta-learners. See appendix \ref{apx:ht_prop_estimator} for details.

\section{Heterogeneous transfer causal effect learners}

Using these building blocks of handling the heterogeneous feature spaces and sharing information about PO functions across domains, we now propose a transfer learning alternative for the standard meta-learners and neural networks (NNs) based CATE estimators, which we refer to as Heterogeneous Transfer Causal Effect (HTCE) learners. See Appendix \ref{apx:implementation_details} for the pseudo-code for the HTCE-learners.

\subsection{Heterogeneous transfer learning for CATE meta-learners} \label{sec:cate_meta_learners}

Consider NN-based implementations for the nuisance functions $\eta = (\mu_0(x), \mu_1(x), \pi(x))$ of each CATE meta-learner. Based on the taxonomy of meta-learners described in \cite{curth2021nonparametric} we divide them into \textit{one-step plug-in learners} and  \textit{two-step learners} and provide HTCE- equivalents for both.

\textbf{One-step plug-in learners.} One-step plug-in learners estimate $\hat{\mu}_1$ and $\hat{\mu}_0$ and then compute CATE as $\hat{\tau}(x) = \hat{\mu}_1(x) - \hat{\mu}_0(x)$. The most common strategies are the T-learner which does not share information between PO functions and the S-learner which does \cite{kunzel2019metalearners}. Here, we consider the simplest S-learner that uses the treatment as an additional feature, while in Section \ref{sec:tarnet} we explore more complex strategies for sharing information between PO functions within a single domain. 

\begin{figure}[h]
    \centering
    \vspace{-0.4cm}
    \subfloat[\textit{HTCE-T-learner}]{\includegraphics[height=2.7cm]{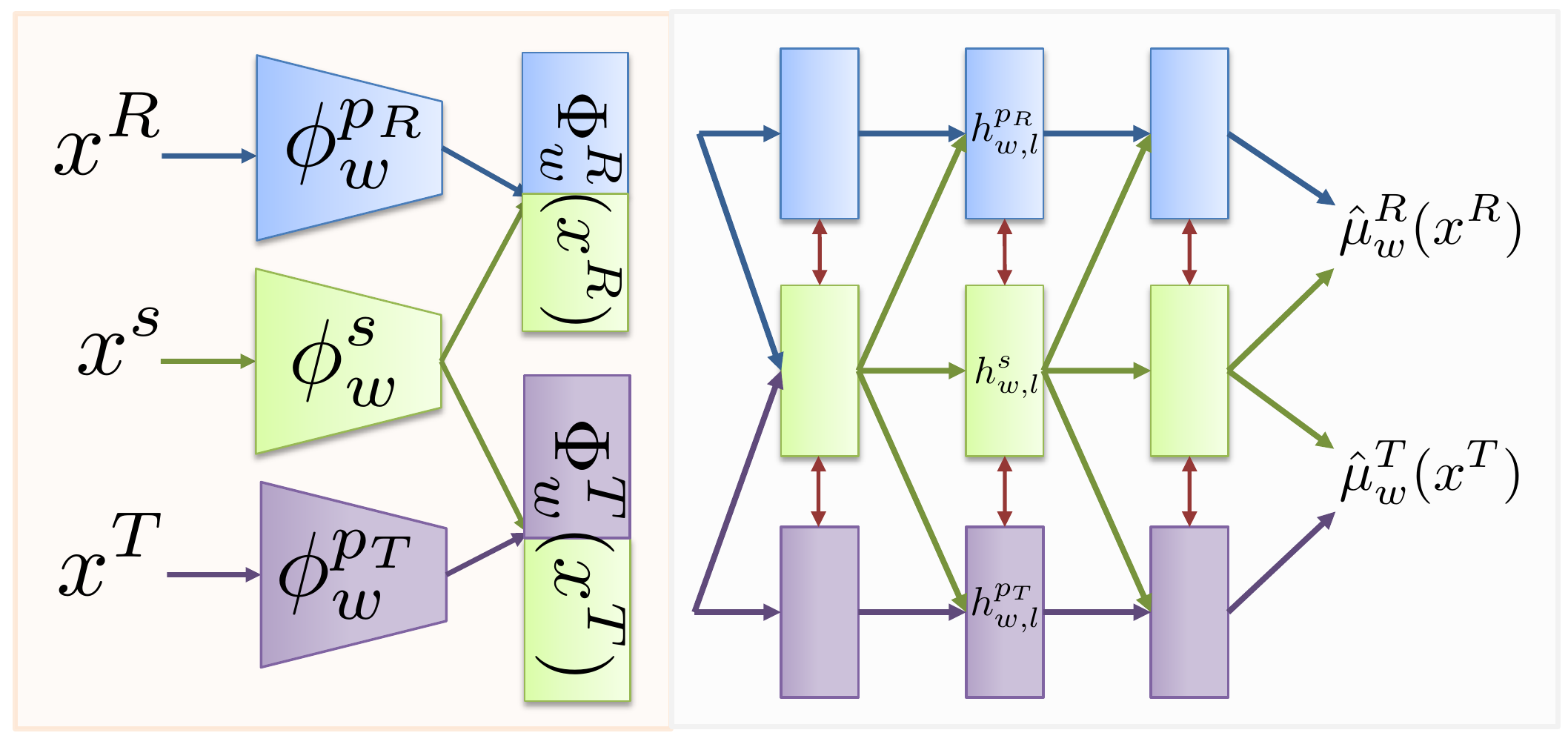}}
    \quad
    \subfloat[\textit{HTCE-S-learner}]{\includegraphics[height=2.7cm]{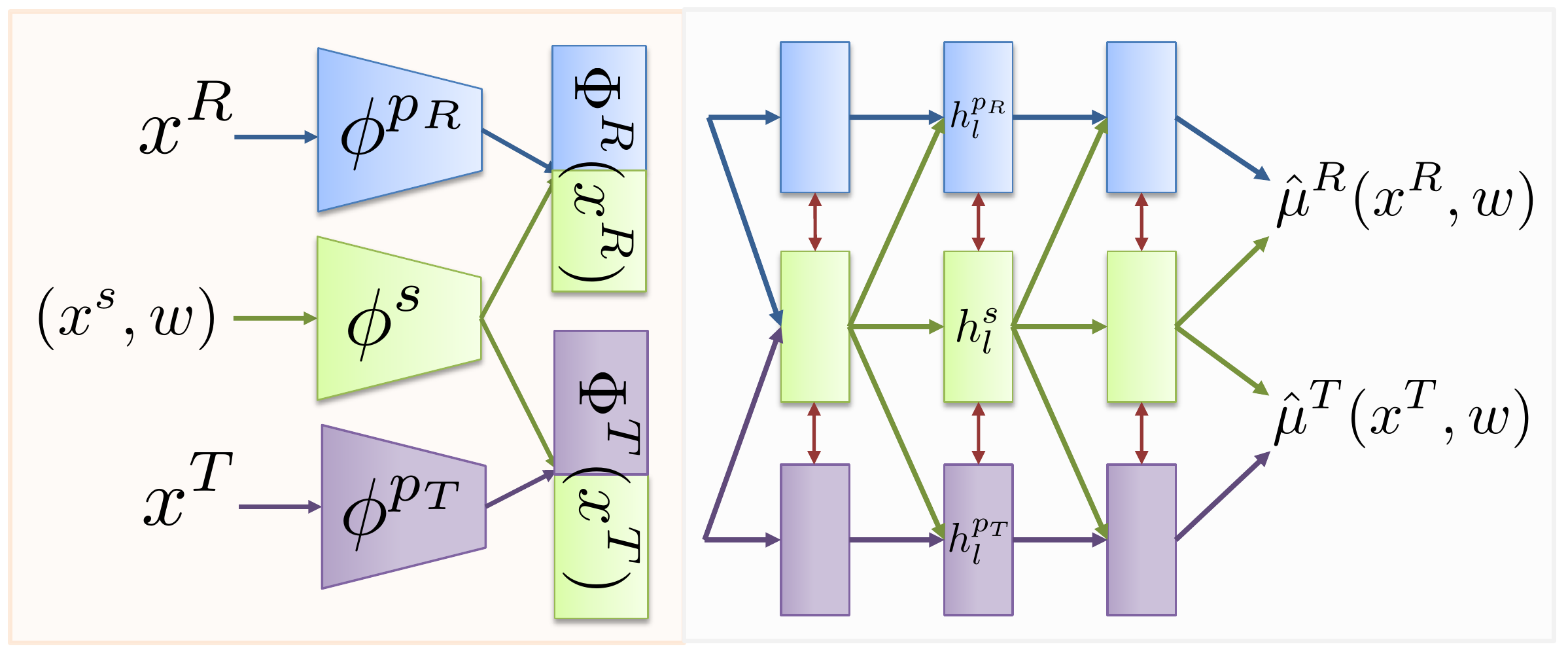} }
    \caption{HTCE-one-step plug-in learners. Notice that the HTCE-T-learner uses treatment specific feature encoders and hypothesis functions in each domain, while the HTCE-S-learner concatenates the treatment to the shared features and shares its components for both treatments in each domain.}
    \label{fig:transfer_learners}
    \vspace{-0.3cm}
\end{figure}

For our \textit{HTCE-T-learner} (Figure \ref{fig:transfer_learners} (a)), we build a model where there is no parameter sharing between $(g_{0}^{R}, g_{1}^{R})$ and between $(g_{0}^{T}, g_{1}^{T})$. We consider treatment-specific encoders $\phi_w^{p_R}, \phi_w^{p_T}, \phi_w^s$ for the heterogeneous feature spaces, such that $\Phi^{R}_w(x^{R}) =  [\phi^{s}_w(x^{s}) ||  \phi^{p_R}_w(x^{R}) ]$ and $\Phi^{T}_w(x^{T}) =  [ \phi^{s}_w(x^{s}) || \phi^{p_T}_w(x^{T})]$. To enable transfer, we use the approach described in Section \ref{sec:sharing_po} to share information between  the hypothesis functions $(g_0^{R}(\Phi^{R}_0(x^{R})), g_0^{T}(\Phi^T_0(x^{T})))$ and $(g_1^{R}(\Phi^{R}_1(x^{R})), g_1^{T}(\Phi^T_1(x^{T})))$. Alternatively, for the \textit{HTCE-S-learner} (Figure \ref{fig:transfer_learners} (b)), we propose concatenating the treatment $w$ to the shared input features $x^s$ and using shared layers to obtain the PO functions in each domain. Thus, we build input representations $\Phi^{R}(x^{R}, w) =  [ \phi^{s}(x^{s}, w) || \phi^{p_R}(x^{R})]$, $\Phi^{T}(x^{T}, w) =  [ \phi^{s}(x^{s}, w) || \phi^{p_T}(x^{T})]$ and we enable transfer by having private and shared layers between the hypothesis functions $g^{R}(\Phi^{R}(x^{R}, w))$ and $g^{T}(\Phi^{T}(x^{T}, w))$ that share parameters for both treatments within a single domain.

\textbf{Two-step learners.} The two-step (direct) learners consist of a first stage that involves obtaining plug-in estimates $\hat{\eta}$ of the nuisance parameters $\eta = (\mu_0, \mu_1, \pi)$ which are used to compute the pseudo-outcome $\tilde{Y}_{\hat{\eta}}$ and a second stage that involves regressing  $\tilde{Y}_{\hat{\eta}}$ on X directly to obtain $\hat{\tau}$. Several strategies based on regression adjustment (RA), propensity weighting (PW) or doubly robust (DR) meta-learner \cite{curth2021nonparametric} have been proposed to compute pseudo-outcomes $\tilde{Y}_{\hat{\eta}}$ which result in unbiased estimates of CATE when $\eta$ is known: $\tau(x) = \mathbb{E}[\tilde{Y}_{\hat{\eta}} \mid X=x]$. These meta-learners have different theoretical properties depending on the sample size and the selection bias present in observational datasets \cite{curth2021nonparametric}, which is why it is important to build a transfer approach to extend all of them.  

\begin{figure}[t]
    \centering
    \includegraphics[width=0.9\textwidth]{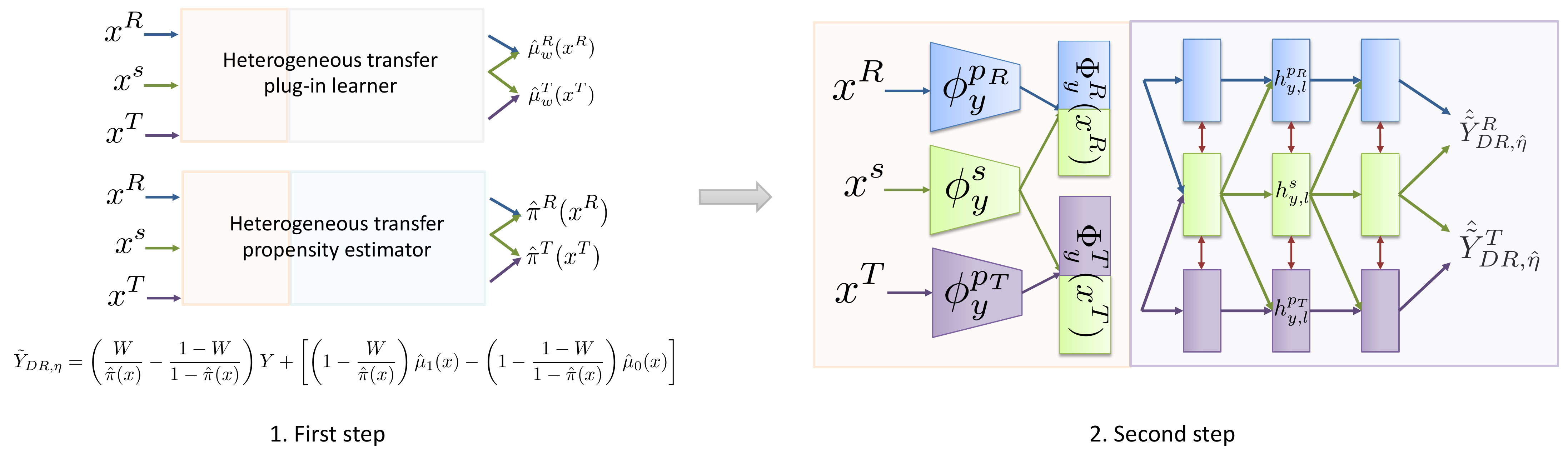}
    \caption{\textit{HTCE-DR-learner}. The first step obtains estimates of $\hat{\eta}$ using our hetereogeneous transfer learning approach that shares information between ($\hat{\mu}_w^R$, $\hat{\mu}_w^T$) and ($\hat{\pi}^R$, $\hat{\pi}^T$), which are then used to compute the pseudo-outcomes $\tilde{Y}_{DR, \eta}$. The second stage regresses the input on the pseud-outcomes and builds a hetereogeneous transfer approach to share information between ($\hat{\tilde{Y}}^R_{DR, \eta}$, $\hat{\tilde{Y}}^T_{DR, \eta}$).}
    \label{fig:transfer_two_step_learners}
    \vspace{-0.2cm}
\end{figure}

We describe here how to build the \textit{HTCE-DR-learner} (Figure \ref{fig:transfer_two_step_learners}) and note that a similar approach can be used for to obtain the  \textit{HTCE-RA-learner}, and the \textit{HTCE-PW-learner} which use a subset of nuissance parameters to compute the pseudo-outcomes. For the first stage of the \textit{HTCE-DR-learner} of estimating the nuisance parameters $\hat{\eta}$, we use an approach similar to the HTCE-T-learner where each nuisance function has its own parameters. To handle the heterogeneous feature spaces, we consider treatment-specific feature encoders for the potential outcome functions $\phi_w^{p_R}, \phi_w^{p_T}, \phi_w^s$ and additional feature encoders for the propensity estimation $\phi_{\pi}^{p_R}, \phi_{\pi}^{p_T}, \phi_{\pi}^s$ such that  $\Phi^{R}_w(x^{R}) =  [ \phi^{s}_w(x^{s}) || \phi^{p_R}_w(x^{R})]$, $\Phi^{T}_w(x^{T}) =  [\phi^{s}_w(x^{s}) || \phi^{p_T}_w(x^{T})]$ and $\Phi^{R}_{\pi}(x^{R}) =  [ \phi^{s}_{\pi}(x^{s}) || \phi^{p_R}_{\pi}(x^{R})]$,  $\Phi^{T}_{\pi}(x^{T}) =  [ \phi^{s}_{\pi}(x^{s}) || \phi^{p_T}_{\pi}(x^{T})]$. Using the approach described in Section \ref{sec:sharing_po} and Appendix \ref{apx:ht_prop_estimator}, we share information across domains between each nuisance function: $(g_w^{R}(\Phi^{R}_w(x^{R})), g_w^{T}(\Phi^T_w(x^{T})))$ and $(g_{\pi}^{R}(\Phi^{R}_{\pi}(x^{R})), g_{\pi}^{T}(\Phi^T_{\pi}(x^{T})))$ to estimate $\hat{\eta}$. We then compute the pseudo-outcomes $\tilde{Y}_{DR, \hat{\eta}}$ as follows \cite{curth2021nonparametric}:
\begin{equation}
   \tilde{Y}_{DR, \eta} =  \left(\frac{W}{\hat{\pi}(x)} - \frac{1-W}{1-\hat{\pi}(x)}\right) Y + \left[\left(1 -\frac{W}{\hat{\pi}(x)}\right)\hat{\mu}_1(x) - \left(1- \frac{1-W}{1-\hat{\pi}(x)}\right)\hat{\mu}_0(x) \right]
\end{equation}

For the second stage of regressing the input on the pseudo-outcomes $\tilde{Y}^{R}_{DR, \hat{\eta}}$, $\tilde{Y}^{T}_{DR, \hat{\eta}}$ directly, we build a similar transfer architecture with feature encoders $\phi_y^{p_R}, \phi_y^{p_T}, \phi_y^s$ to handle the heterogeneous feature, followed by hypothesis functions with private and shared layers $(g_y^{R}(\Phi^{R}_y(x^{R})), g_y^{T}(\Phi^T_y(x^{T})))$, where $\Phi^{R}_y(x^{R}) =  [\phi^{s}_y(x^{s}) || \phi^{p_R}_y(x^{R})]$, $\Phi^{T}_y(x^{T}) =  [\phi^{s}_y(x^{s}) || \phi^{p_T}_y(x^{T})]$.

\subsection{Sharing information between potential outcome functions within each domain} \label{sec:tarnet}

\begin{wrapfigure}{r}{0.55\columnwidth}
	\vspace{-0.6cm}
	\begin{center}
		\includegraphics[width=0.55\columnwidth]{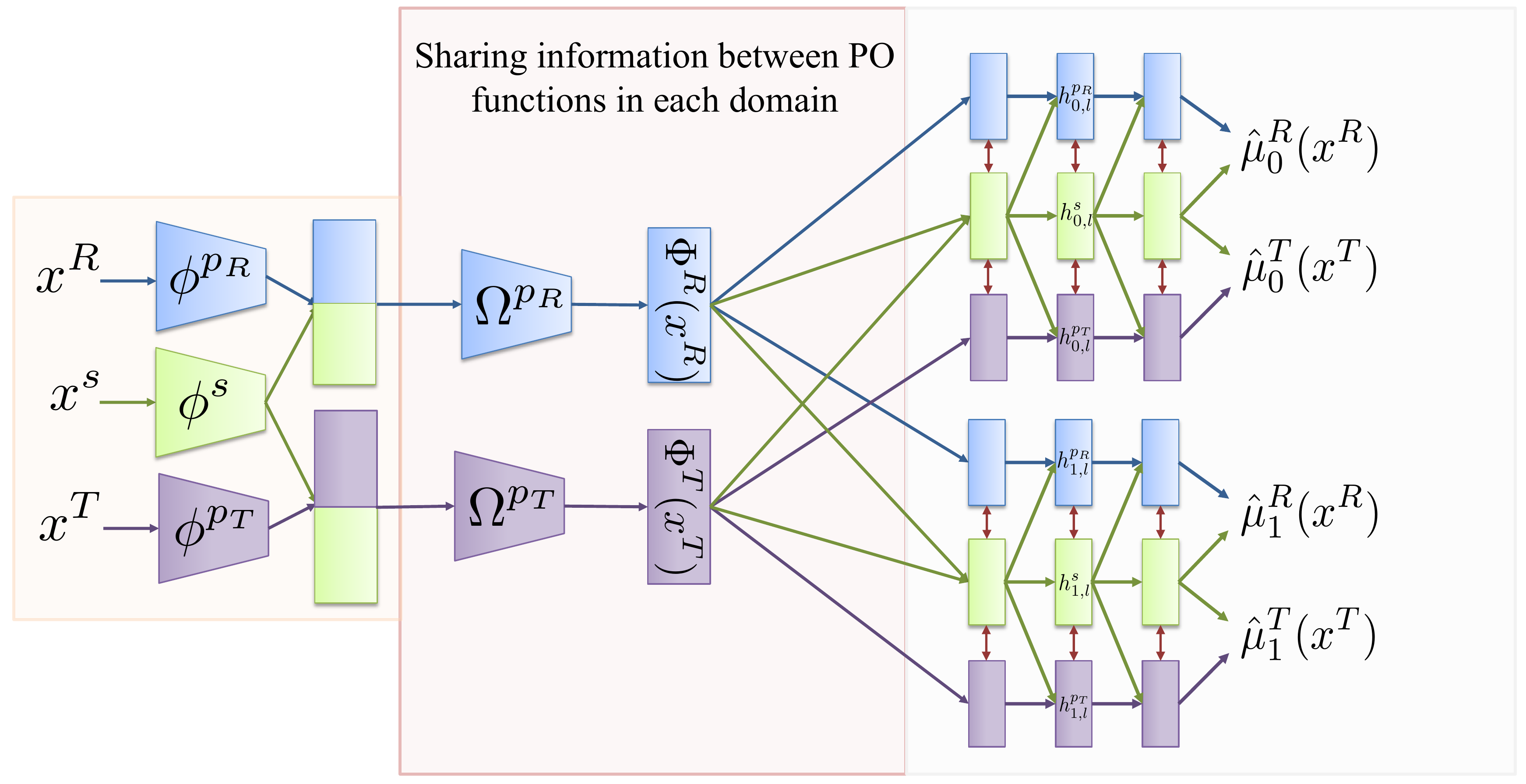}
	\end{center}
	\caption{\textit{HTCE-TARNet.}} 
	\vspace{-0.3cm}
	\label{fig:trasfer_tarnet}
\end{wrapfigure}

In previous sections, we focused on sharing information between PO functions \textit{across} domains $(\mu_1^{R}$, $\mu_1^{T})$ and $(\mu_0^{R}$, $\mu_0^{T})$. However, many methods in the causal inference literature for estimating treatment effects propose different strategies for also sharing information between PO functions \textit{within} a single domain $(\mu_0^{R}$, $\mu_1^{R})$ and $(\mu_0^{T}$, $\mu_1^{T})$ \cite{shalit2017estimating, curth2021nonparametric, hassanpour2019learning}. The most common one is the TARNet architecture \cite{shalit2017estimating} which uses several layers to build a shared representation, followed by outcome-specific layers.

To obtain \textit{HTCE-TARNet} (Figure \ref{fig:trasfer_tarnet}), we propose using feature encoders $\phi^{p_{R}}, \phi^{p_{T}}, \phi^{s}$ that are shared across the different treatments in each domain. Then, we construct another building block that involves adding several representation layers on top of the feature embeddings in each domain using domain-specific encoders $\Omega^{p_T}$ and $\Omega^{p_R}$. These representation layers are domain specific as their purpose is to share information between the PO functions within a single domain. Thus, we obtain $\Phi^{R}(x^{R}) =  \Omega^{p_R}([\phi^{s}(x^{s}) || \phi^{p_R}(x^{R})])$, $\Phi^{T}(x^{T}) = \Omega^{p_T} ([ \phi^{s}(x^{s}) || \phi^{p_T}(x^{T})])$, which are used as input to the corresponding (treatment-specific) hypothesis functions $g^{R}_w(\Phi^{R}(x^{R}))$ and $g^{T}_w(\Phi^{T}(x^{T}))$. 

This approach can be easily extended to recover HTCE- equivalents of other SNet architecture \cite{curth2021nonparametric} that build additional representation layers to share information between PO (and propensity) estimation within each single domain \cite{hassanpour2019learning}. Moreover, several methods have been proposed to account for the selection bias in the observational dataset by building balancing representations \cite{shalit2017estimating, yao2018representation, hassanpour2019learning}. This can also be achieved for our HTCE-learners by enforcing regularization on top of the learnt domain-specific representations: $\Omega^{p_R}(\cdot)$ and $\Omega^{p_T}(\cdot)$. Overall, we have built the HTCE-learners to preserve the characteristics of the different CATE-learners in a single domain, but to handle the heterogeneous feature spaces and to share information between outcome functions across domains.

\section{Evaluation}\label{sec:evaluation}

\subsection{Experimental set-up}

\textbf{Semi-synthetic data generation.} Given that counterfactuals, and subsequently ground truth causal effects are never observed in real-datasets, semi-synthetic data is needed to be able to evaluate methods for estimating CATE. While several benchmarks have been proposed for the standard CATE estimation problem \cite{yoon2018ganite, curth2021nonparametric} in a single domain, none exist for the heterogeneous transfer learning setting. Thus, we propose a new semi-synthetic data simulation to evaluate our HTCE-learners against the benchmarks. Let $X^{R} = (X^{s}, X^{p_R})$ and $X^{T} = (X^{s}, X^{p_T})$ be patient features from real source and target datasets respectively. We simulate outcomes as follows:
\begin{footnotesize}
\begin{align}
    Y^{R}(w) &= \alpha \sum_{j=1}^{D_S} (v_{w, j}^{s} X^{s}_j) / D_S + (1-\alpha) \left[\beta \sum_{j=1}^{D_{p_R}} (v_{w, j}^{p_{R}} X^{p_{R }}_j) / D_{p_R} + (1-\beta)  \sum_{j=1}^{D_{R}} (v_{j}^{{R}} X^{{R}}_j) / D_{R}  \right] + \epsilon_R \\
    Y^{T}(w) &= \alpha \sum_{j=1}^{D_S} (v_{w, j}^{s} X^{s}_j) / D_S + (1-\alpha) \left[\beta \sum_{j=1}^{D_{p_T}} (v_{w, j}^{p_{T}} X^{p_{T}}_j) / D_{p_T} + (1-\beta)  \sum_{j=1}^{D_{T}} (v_{j}^{{T}} X^{{T}}_j) / D_{T}  \right] + \epsilon_T 
\end{align}
\end{footnotesize}%
where $v_{w, j}^{s}, v_{w, j}^{p_{R}}, v_{w, j}^{p_{T}}, v_{j}^{{R}}, v_{j}^{{T}}  \sim \mathcal{N}(-10, 10)$, $\epsilon_R, \epsilon_T \sim \mathcal{N}(0, 0.1)$ and the parameters $\alpha$ and $\beta$ control the amount of sharing between PO across and within domains respectively. For each domain, we simulate the treatment assignment using $W\mid X \sim \text{Bernoulli}(\text{Sigmoid}(\kappa (Y(1) - Y(0))))$, where for $\kappa = 0$ there is no selection bias and the treatments are assigned randomly, while a high $\kappa$ indicates high selection bias. Unless otherwise specified, we set $\alpha=0.5, \beta=0.5$ and $\kappa_{R}=1, \kappa_{T}=1$.   

\textbf{Datasets.}
We perform experiments using patient features from three real datasets with diverse characteristics (e.g. number of features, proportion of categorical features): Twins \cite{almond2005costs}, TCGA \cite{weinstein2013cancer} and MAGGIC \cite{pocock2013predicting}. From the real patient features, the treatment assignments and corresponding outcomes are simulated as described. Refer to Appendix \ref{apx:experimetal_details} for details of the datasets and of the way the source and target domain and the shared and private features are obtained. Due to space limitation, we report here the results on Twins. See Appendix \ref{apx:additional_experiments} for the results on the other datasets.

\textbf{CATE learners and benchmark methods.} The following CATE learners are used in our experiments:  S-Learner, T-Learner, DR-Learner and TARNet. We consider the following benchmarks: (1) training the CATE learners only on the target dataset, (2) using only the shared features between the source and target datasets and (3) using RadialGAN \cite{yoon2018radialgan} to `translate' the source dataset into the target domain. We compare these against our HTCE-learners that leverage the full source and target datasets for training. To ensure a fair comparison, we fixed equivalent hyperparameters in terms of number of layers and units in each layer for the CATE-learners trained on data from the different benchmarks and our HTCE-learners. We describe the hyperarameter used and provide implementation details in Appendix \ref{apx:experimetal_details}. We evaluate the different methods using the
the Root Mean Squared Error of estimating $\tau^{T}(x^T)$, also known as the 
precision of estimating heterogeneous effects (PEHE) \cite{hill2011bayesian}. The results are averaged across 10 runs for which we report the mean PEHE and its standard error.

\newpage
\subsection{Results and discussion} \label{sec:results_and_discussion}

\begin{wraptable}{r}{0.6\columnwidth}
	\centering
	\vspace{-0.4cm}
	\caption{Benchmarks comparison and source of gain analysis in terms of PEHE on Twins dataset.}
	\begin{adjustbox}{max width=0.6\textwidth}
	\begin{tabular}{l|cccc}
			Learner &  S-Learner & T-Learner  & DR-Learner & TARNet  \\
			\midrule
			Target & 0.30 $\pm$ 0.03 &  0.23 $\pm$ 0.02 &  0.20 $\pm$ 0.03 &  0.19 $\pm$ 0.02  \\
			Shared features & 0.46 $\pm$ 0.10 &  0.47 $\pm$ 0.10 &  0.46 $\pm$ 0.09 &  0.46 $\pm$ 0.09   \\
			RadialGAN & 0.21 $\pm$ 0.02 &  0.2 $\pm$ 0.02 &  0.17 $\pm$ 0.01 &  0.19 $\pm$ 0.02   \\
			\midrule
			HTCE - PO sharing & 0.18 $\pm$ 0.03 &  0.15 $\pm$ 0.01 &  0.11 $\pm$ 0.01 &  0.12 $\pm$ 0.01  \\
			HTCE - $\mathcal{L}_{\text{orth}_{z}}$ & 0.14 $\pm$ 0.01 &  0.08 $\pm$ 0.01 & 0.07 $\pm$ 0.01 &  0.11 $\pm$ 0.01  \\
			HTCE - $\mathcal{L}_{\text{ortho}_{\text{PO}}}$ & 0.15 $\pm$ 0.01 &  0.11 $\pm$ 0.01 & 0.10 $\pm$ 0.01 &  0.11 $\pm$ 0.01  \\
			\midrule
			HTCE (ours) &  \textbf{0.12} $\pm$ 0.01 &  \textbf{0.06} $\pm$ 0.01 &  \textbf{0.05} $\pm$ 0.01 &  \textbf{0.09} $\pm$ 0.01 \\
			\bottomrule
	\end{tabular}
	\end{adjustbox}
	\label{tab:benchmars_and_source_of_gain}
	\vspace{-0.5cm}
\end{wraptable}

\textbf{Benchmarks comparison and source of gain}. We first evaluate the HTCE-learners\footnote[1]{The code for the HTCE-learners can be found at \url{https://github.com/vanderschaarlab} and at \url{https://github.com/ioanabica/HTCE-learners}.} against the benchmarks and perform a source of gain analysis to 
evaluate the importance of our different components and loss functions. We evaluate the impact of removing the shared and private layers that enable sharing information between the PO functions across domains (HTCE - PO sharing), removing the orthogonal regularization loss for the learnt shared and private feature representations in Equation \ref{eq:loss_orthogonal_rep} (HTCE - $\mathcal{L}_{\text{orth}_{z}}$) and removing orthogonal regularization loss from the PO layers in Equation \ref{eq:loss_orthogonal_po} (HTCE - $\mathcal{L}_{\text{ortho}_{\text{PO}}}$). Table \ref{tab:benchmars_and_source_of_gain} reports results on the Twins dataset. Note that our HTCE-learners achieve better performance compared to the baselines and that each component we propose brings performance improvements. When analyzing the performance across the different CATE learners we notice that the S-Learner achieves worse performance due to its strong inducting bias of fully-sharing information between PO functions within each domains, which becomes even more restrictive in the transfer setting.

\begin{figure}[H]
    \centering
    \subfloat{\includegraphics[height=2.7cm, clip]{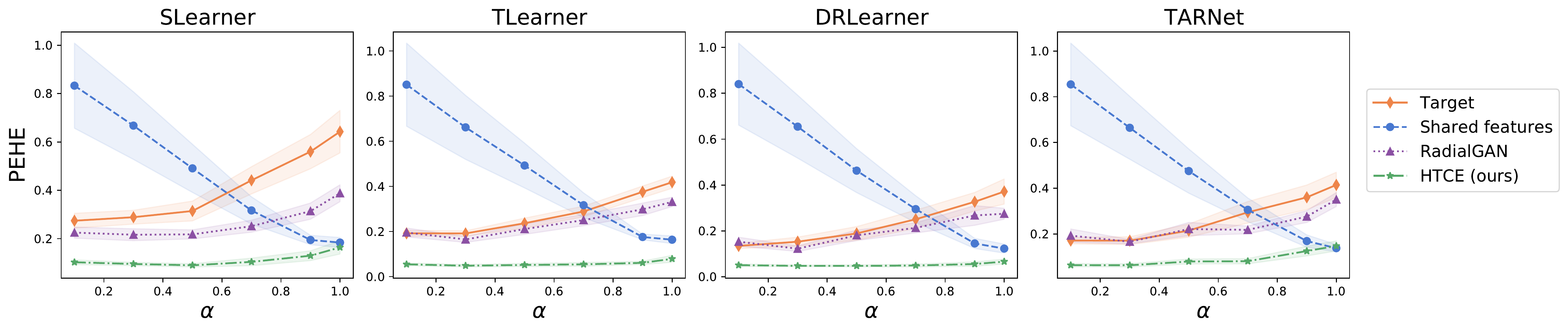}} \vspace{-4mm}
    
    \subfloat{\includegraphics[height=2.7cm, clip]{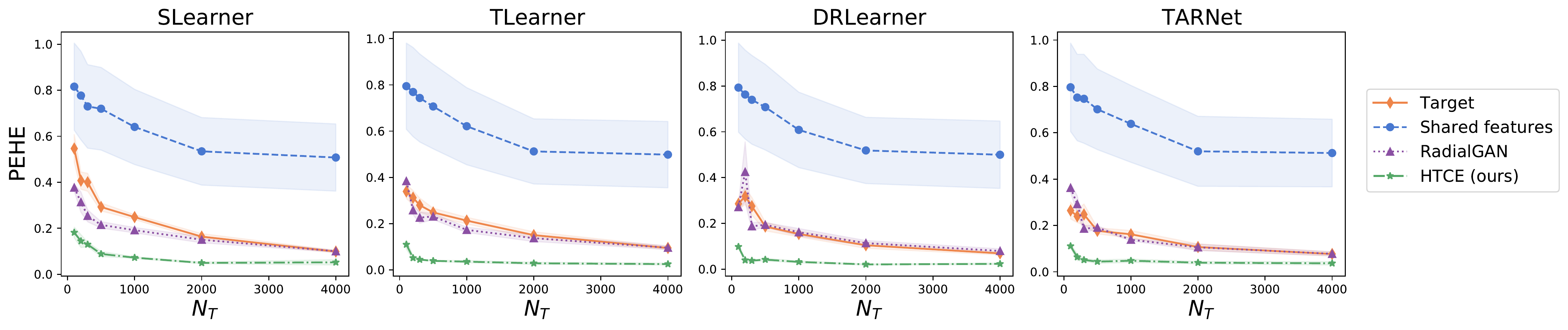}}
    \caption{Performance comparison on Twins when varying the information sharing between PO functions across domains through $\alpha$ (top) and the size of the target dataset through $N_T$ (bottom).}
    \label{fig:performance_alpha_size_target}
\end{figure}

\textbf{Varying the information sharing between domains}. To further gain insights into the differences between the benchmarks under different settings, we vary the parameter $\alpha$, which controls the amount of information shared between the PO in the source and target dataset. Figure \ref{fig:performance_alpha_size_target} (top) reports the results for this analysis on the Twins dataset.  
Note that our flexible approach for information sharing between PO functions across domains achieves good performance both when the PO functions are significantly different between the source and target dataset ($\alpha=0.1$) and also when they have the same functional form  and only depend on the shared features ($\alpha=1.0$). As a sanity check, we also notice that for  ($\alpha=1.0$) the performance of the HTCE-learners matches the one of the corresponding CATE learners trained only on the shared features between the two domains. Moreover, note that for this setting of $\alpha=1.0$ when the PO only depend on the shared features between the two domains, the different CATE learners trained only on the target dataset have a decrease in performance as they also need to perform the task of implicit feature selection with limited amount of data.

\textbf{Varying the target dataset size.} We also evaluate performance when varying the size of the target dataset $N_T$. We report in Figure \ref{fig:performance_alpha_size_target} (bottom) the results on the Twins datasets. Note that while our HTCE-learners still bring benefits for all values of $N_T$, the most performance improvements are when the target dataset size is small and we get diminishing returns as $N_T$ increases. Moreover, note that TARNet trained only on the target dataset reaches the performance of the HTCE-TARNet learner with less amount of data compered to T-learner trained on the target dataset vs. the  HTCE-T-learner. 

\begin{wrapfigure}{r}{0.6\columnwidth}
	\vspace{-0.1cm}
	\begin{center}
		\includegraphics[width=0.6\columnwidth]{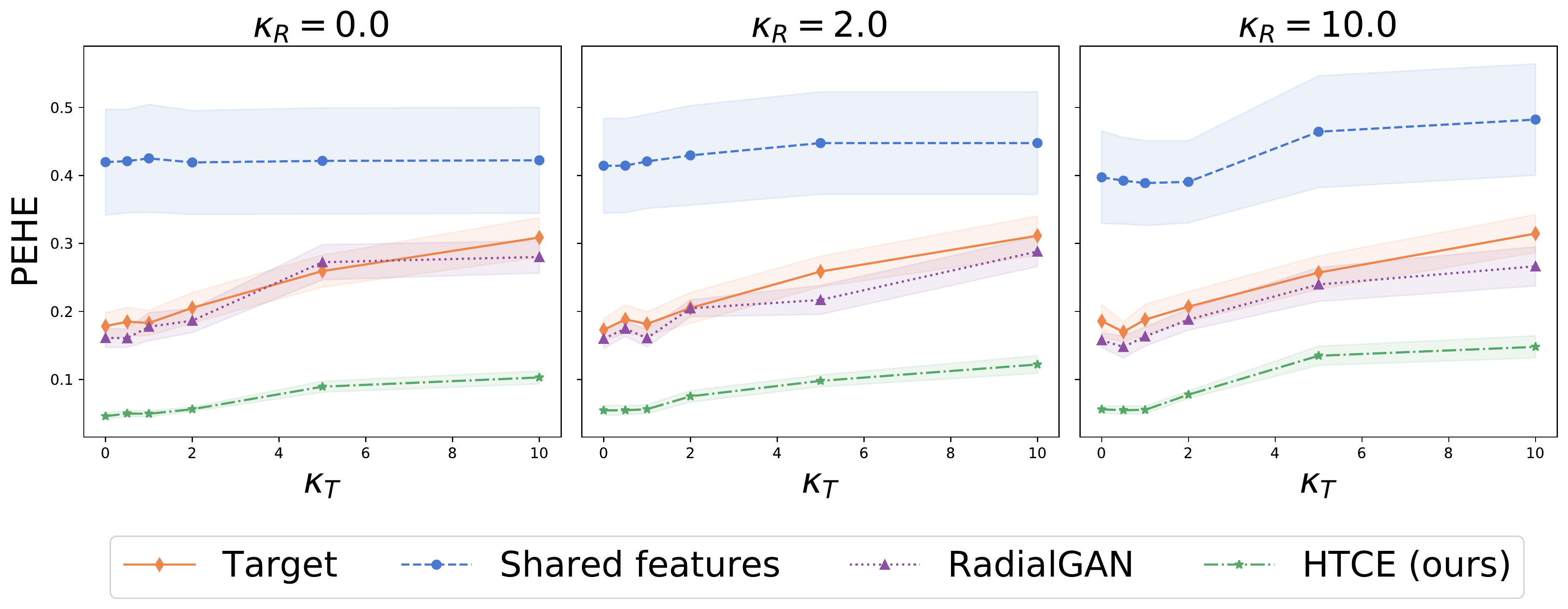}
	\end{center}
	\vspace{-0.1cm}
	\caption{Performance comparison on Twins for DR-learner when varying the selection bias $\kappa_R$ and $\kappa_T$ in the source and target datasets.}
	\vspace{-0.1cm}
	\label{fig:performance_selection_bias}
\end{wrapfigure}

\textbf{Effect of selection bias.} Finally, we investigate how the selection bias in the source and target datasets impact transfer performance. We consider three setting for the selection bias in the source dataset $\kappa_R = 0.0$ for random treatment assignment, $\kappa_R = 2.0$ for moderate and $\kappa_R = 10.0$ for strong selection bias. For each setting of $\kappa_R$, we vary the selection bias in the target dataset $\kappa_T$ from $\kappa_T = 0.0$ to $\kappa_T = 10.0$ and report the results on Twins for the DR-learner in Figure \ref{fig:performance_selection_bias}. See Appendix \ref{apx:additional_experiments} for results on the other learners and datasets. Note that our HTCE-DR-learner has consistent performance when increasing the selection bias in both datasets and when there are significant discrepancies in the treatment assignment mechanism between the source and target datasets.

\section{Discussion} \label{sec:discussion}

We proposed heterogeneous transfer causal effect (HTCE) learners capable of improving the estimation of CATE on a target domain by leveraging data from a related source domain. This represents the first work that addresses the heterogeneous transfer learning problem for CATE estimation which is prevalent in many causal inference application areas where the personalized effects of interventions for a new population need to be estimated from small target datasets with different features than the available source datasets \cite{wiens2014study, forney2021causal}. We hope that the insights gained from this paper, in terms of the need for sharing information between PO functions to enable transfer 
and the differences between CATE learners in the transfer setting, will help guide further model development for this problem.

In terms of limitations and directions for future work, while we focused on the binary treatment setting which is the most common in the causal inference literature, addressing the problem of heterogeneous transfer learning is equally important for more complex treatment settings such as treatments with dosage \cite{bica2020scigan, schwab2019learning}, treatment combinations \cite{qian2021estimating, parbhoo2021ncore} and treatments in the temporal setting \cite{lim2018forecasting, bica2020crn, bica2020time, berrevoets2021disentangled, qian2021synctwin}. Moreover, while we demonstrate the benefits of our HTCE-learners experimentally, future work should also provide theoretical guarantees on their performance in terms of the similarities between the source and target datasets. We provide further discussion of possible extensions and limitations of our method in Appendix \ref{apx:additional_extensions}. Finally, we acknowledge that in areas such as healthcare, acting based on incorrect estimates from such causal inference models can have life-threatening implications. Because of this, it becomes crucial that such models undergo extensive testing through clinical trials and expert validation and that they are only used for decision support alongside clinicians.

\begin{ack}
We would like to thank the reviewers for their valuable feedback. Moreover, we would also like to thank Alicia Curth for insightful discussions. The research presented in this paper was supported by The Alan Turing Institute, under the EPSRC grant EP/N510129/1, by the US Office of Naval Research (ONR), and by the National Science Foundation (NSF) under grant number 1722516. 
\end{ack}

\newpage
\bibliography{refs.bib}
\bibliographystyle{unsrt}

\newpage

\appendix
\section{Expanded related works} \label{apx:related_works}

In this section, we further expand on the related works described in Section \ref{sec:related_works}.

\textbf{Additional methods that leverage multiple datasets for causal inference.} While several other methods propose leveraging multiple datasets to improve CATE estimation, these focus on combining observational data with randomized data from clinical trials with the aim of handling the bias from hidden confounders \cite{kallus2018removing, ilse2021combining, hatt2022combining}. Moreover, all of these methods learn potential outcome functions that are shared between the different domains. Conversely, we assume that there are no hidden confounders and propose an approach that can share information from a source domain to improve CATE estimation on a target domain of interest. 

 A different strand of work aims to address the problem of identifiability of causal effects for an observational dataset of interest by leveraging data from other datasets. In particular, \cite{pearl2011transportability} investigates how to transfer average treatment (causal) effects obtained from experimental data (such as randomized clinical trials) to an observed population that may have different distribution of covariates, treatments and outcomes and where the causal relationship of interest cannot be identified using only the observational data. The paper assesses under which conditions such average causal effects can be transported according to the differences between the randomized and observational data. The authors also provide a brief discussion of how to transfer these average causal estimates between observational datasets. Alternatively, \cite{bareinboim2016causal} aims to help identify the average effects of interventions on a target population of interest by integrating multiple types of auxiliary data: data from a randomized study on the same population, data from an observational study on the same population, selection biased data from the same population and data from a randomized study from a different population. All of the auxiliary datasets have the same set of features as the target dataset. Both \cite{pearl2011transportability, bareinboim2016causal} consider this transportability problem of average treatment effects in the context of causal diagrams. This is different from our set-up for CATE estimation where we assume that the potential outcomes are identifiable in both the source and target domains. Moreover, we also handle the case of heterogeneous feature spaces and only assume access to a source domain larger than a target domain, without putting any restrictions on whether this data is experimental or observational.

\textbf{Transfer learning and domains adaptation in the predictive setting}. Another method loosely related to ours is the one of \cite{moon2017completely},  which considers the case of having both heterogeneous feature and label spaces in the context of natural language processing and proposes a method that learns a common embedding for the features and labels and then builds a mapping between them. In this paper, we consider a shared label space.  In addition, \cite{magliacane2018domain} uses a causal approach to address the problem of domain adaptation for the predictive setting, where it considers labelled data in one or more source domains, unlabelled data in the target domain, same feature spaces between the source and target domains and aims to learn predictive functions that are invariant to the changes between domains to be able to reliably estimate the outcomes in the target domain. \cite{magliacane2018domain} proposes tackling this unsupervised domain adaptation by modelling the different distributions between the source and target domains as different contexts of a single underlying system (where the context represents interventions causing the distribution shifts between domains, such that the source and target domains come from different interventions). 

Table \ref{tab:related-works} provides a comparison of our method with the most relevant related works. Note that our heterogeneous transfer learning approach is designed to share information between one source domain and a target domains. Moreover, we assume access to labels on the target domain and build target-specific outcome functions that share a common structure with the source domain. 

\newcolumntype{A}{>{\centering\arraybackslash}m{0.9 cm}}
\begin{table*}[h]
\begin{center}
\begin{small}
\setlength\tabcolsep{1pt}
\begin{adjustbox}{max width=\textwidth}
\begin{tabular}{Alccccccc} 
\toprule
& Method & \makecell{Number of \\ domains} & \makecell{Heterogeneous \\ feature spaces} & \makecell{Heterogeneous \\ label spaces} & \makecell{Access to labels \\ on target domain}  & \makecell{Domain-specific \\ outcome functions} \\
\midrule
\multirow{3}{*}{\rotatebox[origin=c]{90}{\scalebox{0.8}{\makecell{Causal \\Inference}}}}
& CATE learners \cite{curth2021nonparametric, shalit2017estimating, curth2021inductive} & 1 & No & No & No & No   \\
&  Johansson et al. \cite{johansson2018learning} & 2 & No & No & No & No  \\
& Shi et al. \cite{shi2021invariant} & $N$ & No & No & Yes & No  \\
& Kunzel et al. \cite{kunzel2018transfer}& 2 & No & No & Yes & Yes \\
\midrule
\multirow{3}{*}{\rotatebox[origin=c]{90}{\scalebox{0.8}{\makecell{Learning ac- \\ ross domains}}}}
& Ganin et al. \cite{ganin2016domain}  & 2 & No & No & No & No  \\
&  Magliacane et al. \cite{magliacane2018domain} & N &  No &  No & No &  No  \\
& Rudner et al. \cite{ruder2019latent}  & $N$ & No & No & Yes & Yes  \\
&  Yoon et al. \cite{yoon2018radialgan} & $N$ & Yes & No & Yes  & Yes  \\
&  Moon et al. \cite{moon2017completely} & 2 & Yes & Yes & Yes & No  \\
\midrule
 \multicolumn{2}{l}{~~(Ours) HTCE-learners} & 2 & Yes & No & Yes  &  Yes  \\
\bottomrule
\end{tabular}
\end{adjustbox}
\end{small}
\end{center}
\vspace{-0.75em}
\caption{Comparison of our proposed method with related works.}
\label{tab:related-works}
\vspace{-2mm}
\end{table*}

\newpage
\section{Heterogeneous transfer propensity estimator} \label{apx:ht_prop_estimator}

\begin{figure}[h]
    \centering
    \includegraphics[width=0.5\textwidth]{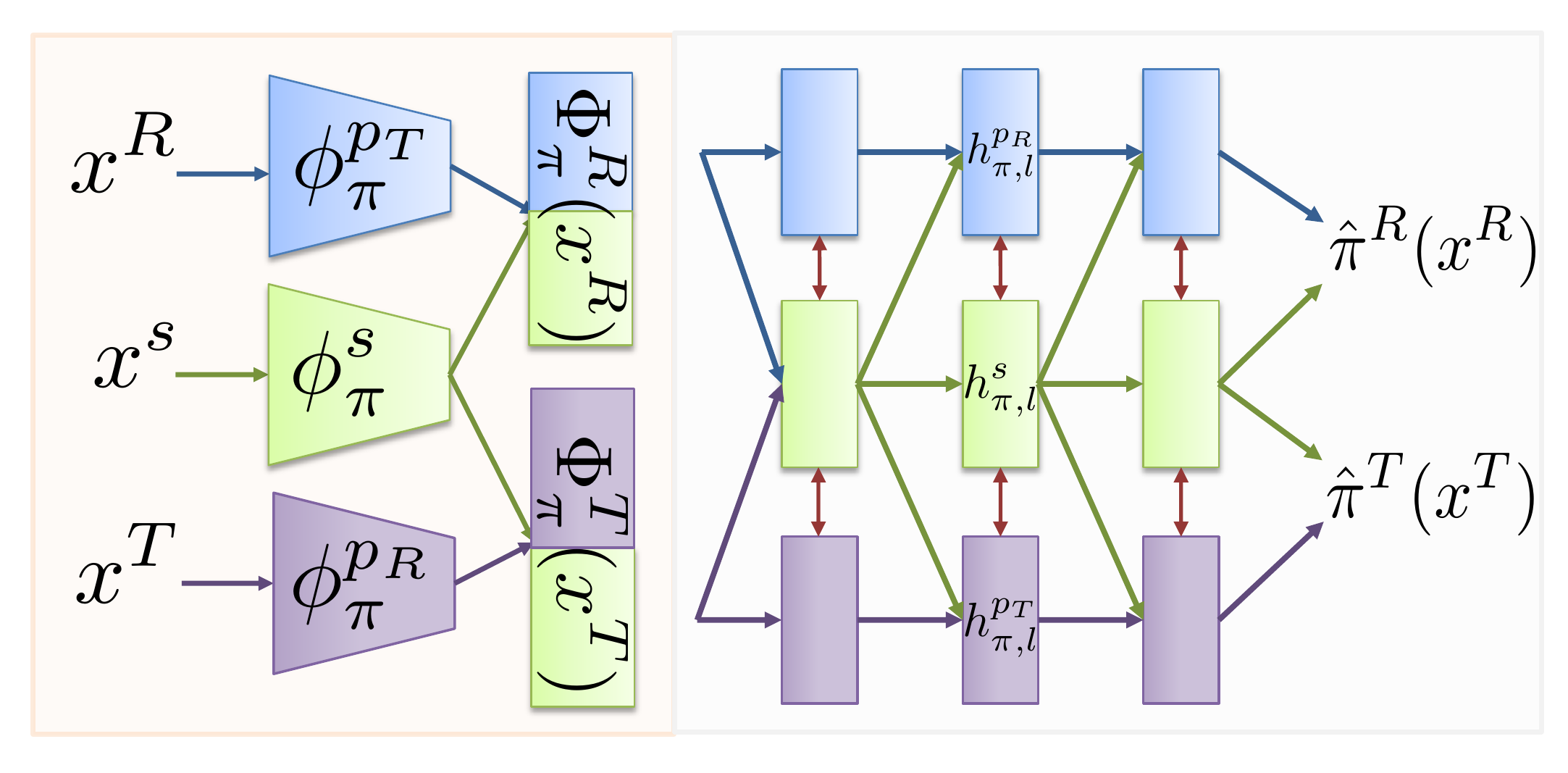}
    \caption{HTCE-propensity estimator.}
    \label{fig:htce_propensity}
\end{figure}

For our \textit{HTCE-propensity estimator} that is used as part of the two-step meta-learners (e.g. HTCE-DR-learner) we build a model that shares information between the treatment assignment mechanisms in the source and target datasets as illustrated in Figure \ref{fig:htce_propensity}. In particular, our HTCE-propensity estimator consists of encoders $\phi_{\pi}^{p_R}, \phi_{\pi}^{p_T}, \phi_{\pi}^s$ for the heterogeneous feature spaces, such that $\Phi^{R}_{\pi}(x^{R}) =  [\phi^{s}_{\pi}(x^{s}) ||  \phi^{p_R}_{\pi}(x^{R}) ]$ and $\Phi^{T}_{\pi}(x^{T}) =  [ \phi^{s}_{\pi}(x^{s}) || \phi^{p_T}_{\pi}(x^{T})]$. To enable transfer, we use an approach similar to the one described in Section \ref{sec:sharing_po}, but this time to share information between the hypothesis functions for the propensity estimation $(g_{\pi}^{R}(\Phi^{R}_{\pi}(x^{R})), g_{\pi}^{T}(\Phi^T_{\pi}(x^{T})))$. Moreover, we use the following outcome loss:
\begin{equation}
    \mathcal{L}_{\pi} = \sum_{i=1}^{N_{R}} l(w_i, g_{\pi}^{R}(\Phi^R(x_i^{R}))) + \sum_{i=1}^{N_{T}} l(w_i, g_{\pi}^{T}(\Phi^T(x_i^{T})))
\end{equation}
where $l(\cdot, \cdot)$ represents the binary cross-entropy loss (for binary treatment).

\section{Pseudo-code for the HTCE-learners} \label{apx:implementation_details}

Algorithm \ref{alg:htce_tlearner} provides the pseudo-code for training our proposed HTCE-T-learner and Algorithm \ref{alg:htce_tarnet} provides the pseudo-code for training our proposed HTCE-TARNet. Note that a similar training procedure can be used for the HTCE-S-Learner, HTCE-propensity estimator and in the second stage of the two-step learners (e.g. HTCE-DR-learner). The main differences are in the encoder used, in terms of sharing layers between PO functions within each domain and the outcome used for computing the corresponding loss function. Refer to Section \ref{sec:cate_meta_learners} and Appendix \ref{apx:ht_prop_estimator} for architectural details. Similarly to \cite{curth2021nonparametric}, we do not use cross-fitting for the two-step learners (HTCE-DR-learner). Refer to \cite{curth2021nonparametric} for more details. For simplicity, we set the mini-batch size to be the same in both the source and target datasets: $B_T=B_R=B$. Moreover, note that in practice, we multiply $\mathcal{L}_{\text{orth}_{z}}$ and $\mathcal{L}_{\text{orth}_{\text{PO}}}$ by $0.01$ to ensure that all losses have a similar scale and we use early stopping on the validation dataset from the target domain to check for convergence. 

\newpage
\begin{algorithm}[H]
\begin{algorithmic}[1] 
\State Input: Dataset with expert demonstrations from source $\mathcal{D}^R$ and target  $\mathcal{D}^T$ domains, learning rate $\lambda$, mini-batch size for source dataset $B_R$, mini-batch size for target dataset $B_T$
\State Initialize: $\theta$ (all parameters of the HTCE-T-Learner) 
\While{not converged}
    \State Sample mini-batch of $B_R$ demonstrations from the source dataset $(x^R_i, w_i, y_{i}) \sim \mathcal{D}^{R}$ and a minibatch of $B_T$ demonstrations from the target dataset  $(x^T_i, w_i, y_{i}) \sim \mathcal{D}^{T}$. 
    \State \textbf{Handle heterogeneous feature spaces}:
    \For{$i=1\dots B_R$} \Comment{Process batch from source domain.}
    \State $z_i^{s_R} = \phi_{w_i}^{s}(x_i^s)$, where $x_i^s$ is part of $x_i^{R}$; $z_i^{p_R} = \phi_{w_i}^{R}(x_i^R)$
    \EndFor
    \State Set: $\zeta^{s_R} = [z_1^{s_R} \dots z_{B_R}^{s_R}]^{\top}$, $\zeta^{p_R} = [z_1^{p_R} \dots z_{B_R}^{p_R}]^{\top}$ 
    \For{$i=1\dots B_T$} \Comment{Process batch from target domain.}
    \State $z_i^{s_T} = \phi_{w_i}^{s}(x_i^s)$, where $x_i^s$ is part of $x_i^{T}$; $z_i^{p_T} = \phi_{w_i}^{R}(x_i^T)$
    \EndFor
    \State Set: $\zeta^{s_T} = [z_1^{s_T} \dots z_{B_T}^{s_T}]^{\top}$, $\zeta^{p_T} = [z_1^{p_T} \dots z_{B_T}^{p_T}]^T$ 
    \State Compute orthogonal regularization loss for the feature representations:
    \begin{equation*}
    \mathcal{L}_{\text{orth}_{z}} =  \| {\zeta^{s_R}}^{\top} \zeta^{p_R} \|^{2}_{F}  + \| {\zeta^{s_T}}^{\top} \zeta^{p_{T}} \|^{2}_{F}
    \end{equation*}
    \Statex
    \State \textbf{Share information between PO functions across domains}:
    \For{$i=1\dots B_R$} \Comment{Process batch from source domain.}
    \For{$l=1\dots L$}
    \If{$l==1$}
    \State $\tilde{h}_{w_i, 1}^{p_R} = [z_i^{s_R} || z_i^{p_R}]$; $\tilde{h}_{w_i, 1}^{s} = \tilde{h}_{w_i, 1}^{p_R}$
    \Else 
    \State $\tilde{h}_{w_i, l}^{p_R} = [ h_{w_i, l-1}^{s} || h_{w_i, l-1}^{p_R}]$; $\tilde{h}_{w, l}^{s} = [h_{w, l-1}^{s}]$.
    \EndIf
    \State $h_{w_i, l}^{s} = \text{Shared\_Layer}_{w_i,l }(\tilde{h}_{w_i, l}^{s})$;  $h_{w_i, l}^{p_R} = \text{Private\_Layer}^R_{w_i,l }(\tilde{h}_{w_i, l}^{p_R})$
    \EndFor
    \State  $\hat{y}_i^{R} = \psi(h_{w_i, L}^{p_R} + h_{w_i, L}^{s})$ 
    \EndFor
    \For{$i=1\dots B_T$} \Comment{Process batch from target domain.}
    \For{$l=1\dots L$}
    \If{$l==1$}
    \State $\tilde{h}_{w_i, 1}^{p_T} = [z_i^{s_T} || z_i^{p_T}]$; $\tilde{h}_{w_i, 1}^{s} = \tilde{h}_{w_i, 1}^{p_T}$ 
    \Else 
    \State $\tilde{h}_{w_i, l}^{p_T} = [h_{w_i, l-1}^{s} || h_{w_i, l-1}^{p_T}]$, $\tilde{h}_{w_i, l}^{s} = [h_{w_i, l-1}^{s}]$.
    \EndIf
    \State $h_{w_i, l}^{s} = \text{Shared\_Layer}_{w_i,l }(\tilde{h}_{w_i, l}^{s})$; $h_{w_i, l}^{p_T} = \text{Private\_Layer}^T_{w_i,l }(\tilde{h}_{w_i, l}^{p_T})$
    \EndFor
    \State  $\hat{y}_i^{T} = \psi(h_{w_i, L}^{p_T} + h_{w_i, L}^{s})$
    \EndFor
    \State Compute outcome loss: 
    \begin{equation*}
    \mathcal{L}_y = \sum_{i=1}^{B_R} l(y_i, \hat{y}_i^{R}) + \sum_{i=1}^{B_T} l(y_i, \hat{y}_i^{T})
    \end{equation*}
    \State Compute orthogonal regularization loss for the PO shared and private layers: 
    \begin{equation*}\mathcal{L}_{\text{orth}_{\text{PO}}} = \sum_{w\in \{0, 1\}} \sum_{l=1}^{L}  \| {\theta_{w, l}^{s}}^{\top} \theta^{p_R}_{w, l, 1:m_{l-1}^s} \|^{2}_{F}  + \| {\theta_{w, l}^{s}}^{\top} \theta^{p_T}_{w, l, 1:m_{w, l-1}^s} \|^{2}_{F}
    \end{equation*}
    \Statex
    \State Parameters update: 
    \State $\theta \leftarrow \theta - \lambda \nabla_{\theta} (\mathcal{L}_{y} + \mathcal{L}_{\text{orth}_{z}} + \mathcal{L}_{\text{orth}_{\text{PO}}})$
\EndWhile
\State \textbf{Output:} Learnt parameters $\theta$ 
\caption{Pseudo-code for training the HTCE-T-learner.} \label{alg:htce_tlearner}
\end{algorithmic} 
\end{algorithm}

\begin{algorithm}[H]
\begin{algorithmic}[1] 
\State Input: Dataset with expert demonstrations from source $\mathcal{D}^R$ and target  $\mathcal{D}^T$ domains, learning rate $\lambda$, mini-batch size for source dataset $B_R$, mini-batch size for target dataset $B_T$
\State Initialize: $\theta$ (all parameters of the HTCE-TARNet) 
\While{not converged}
    \State Sample mini-batch of $B_R$ demonstrations from the source dataset $(x^R_i, w_i, y_{i}) \sim \mathcal{D}^{R}$ and a minibatch of $B_T$ demonstrations from the target dataset  $(x^T_i, w_i, y_{i}) \sim \mathcal{D}^{T}$. 
    \State \textbf{Handle heterogeneous feature spaces}:
    \For{$i=1\dots B_R$} \Comment{Process batch from source domain.}
    \State $z_i^{s_R} = \phi^{s}(x_i^s)$, where $x_i^s$ is part of $x_i^{R}$; $z_i^{p_R} = \phi^{R}(x_i^R)$
    \EndFor
    \State Set: $\zeta^{s_R} = [z_1^{s_R} \dots z_{B_R}^{s_R}]^{\top}$, $\zeta^{p_R} = [z_1^{p_R} \dots z_{B_R}^{p_R}]^{\top}$ 
    \For{$i=1\dots B_T$} \Comment{Process batch from target domain.}
    \State $z_i^{s_T} = \phi^{s}(x_i^s)$, where $x_i^s$ is part of $x_i^{T}$; $z_i^{p_T} = \phi^{R}(x_i^T)$
    \EndFor
    \State Set: $\zeta^{s_T} = [z_1^{s_T} \dots z_{B_T}^{s_T}]^{\top}$, $\zeta^{p_T} = [z_1^{p_T} \dots z_{B_T}^{p_T}]^T$ 
    \State Compute orthogonal regularization loss for the feature representation:
    \begin{equation*}
    \mathcal{L}_{\text{orth}_{z}} =  \| {\zeta^{s_R}}^{\top} \zeta^{p_R} \|^{2}_{F}  + \| {\zeta^{s_T}}^{\top} \zeta^{p_{T}} \|^{2}_{F}
    \end{equation*}
    \Statex
    \State \textbf{Share information between PO functions within each domain}:
    \For{$i=1\dots B_R$} \Comment{Process batch from source domain.}
    \State $\tilde{z}_i^{R} = \Omega^{p_R}([z_i^{s_R} || z_i^{p_R}])$
    \EndFor
    \For{$i=1\dots B_T$} \Comment{Process batch from target domain.}
    \State $\tilde{z}_i^{T} = \Omega^{p_T}([z_i^{s_T} || z_i^{p_T}])$
    \EndFor
    \Statex
    \State \textbf{Share information between PO functions across domains}:
    \For{$i=1\dots B_R$} \Comment{Process batch from source domain.}
    \For{$l=1\dots L$}
    \If{$l==1$}
    \State $\tilde{h}_{w_i, 1}^{p_R} = \tilde{z}_i^{R}$; $\tilde{h}_{w_i, 1}^{s} = \tilde{h}_{w_i, 1}^{p_R}$
    \Else 
    \State $\tilde{h}_{w_i, l}^{p_R} = [ h_{w_i, l-1}^{s} || h_{w_i, l-1}^{p_R}]$; $\tilde{h}_{w, l}^{s} = [h_{w, l-1}^{s}]$.
    \EndIf
    \State $h_{w_i, l}^{s} = \text{Shared\_Layer}_{w_i,l }(\tilde{h}_{w_i, l}^{s})$;  $h_{w_i, l}^{p_R} = \text{Private\_Layer}^R_{w_i,l }(\tilde{h}_{w_i, l}^{p_R})$
    \EndFor
    \State  $\hat{y}_i^{R} = \psi(h_{w_i, L}^{p_R} + h_{w_i, L}^{s})$ 
    \EndFor
    \For{$i=1\dots B_T$} \Comment{Process batch from target domain.}
    \For{$l=1\dots L$}
    \If{$l==1$}
    \State $\tilde{h}_{w_i, 1}^{p_T} = \tilde{z}_i^{T}$; $\tilde{h}_{w_i, 1}^{s} = \tilde{h}_{w_i, 1}^{p_T}$ 
    \Else 
    \State $\tilde{h}_{w_i, l}^{p_T} = [h_{w_i, l-1}^{s} || h_{w_i, l-1}^{p_T}]$, $\tilde{h}_{w_i, l}^{s} = [h_{w_i, l-1}^{s}]$.
    \EndIf
    \State $h_{w_i, l}^{s} = \text{Shared\_Layer}_{w_i,l }(\tilde{h}_{w_i, l}^{s})$; $h_{w_i, l}^{p_T} = \text{Private\_Layer}^T_{w_i,l }(\tilde{h}_{w_i, l}^{p_T})$
    \EndFor
    \State  $\hat{y}_i^{T} = \psi(h_{w_i, L}^{p_T} + h_{w_i, L}^{s})$
    \EndFor
    \State Compute outcome loss: 
    \begin{equation*}
    \mathcal{L}_y = \sum_{i=1}^{B_R} l(y_i, \hat{y}_i^{R}) + \sum_{i=1}^{B_T} l(y_i, \hat{y}_i^{T})
    \end{equation*}
    \State Compute orthogonal regularization loss for the PO shared and private layers: 
    \begin{equation*}\mathcal{L}_{\text{orth}_{\text{PO}}} = \sum_{w\in \{0, 1\}} \sum_{l=1}^{L}  \| {\theta_{w, l}^{s}}^{\top} \theta^{p_R}_{w, l, 1:m_{l-1}^s} \|^{2}_{F}  + \| {\theta_{w, l}^{s}}^{\top} \theta^{p_T}_{w, l, 1:m_{w, l-1}^s} \|^{2}_{F}
    \end{equation*}
    \Statex
    \State Parameters update: 
    \State $\theta \leftarrow \theta - \lambda \nabla_{\theta} (\mathcal{L}_{y} + \mathcal{L}_{\text{orth}_{z}} + \mathcal{L}_{\text{orth}_{\text{PO}}})$
\EndWhile
\State \textbf{Output:} Learnt parameters $\theta$ 
\caption{Pseudo-code for training the HTCE-TARNet.} 
\label{alg:htce_tarnet}
\end{algorithmic} 
\end{algorithm}

\section{Experimental details} \label{apx:experimetal_details}

\subsection{Dataset description}

In this section, we provide a description of the three datasets used for evaluation: MAGGIC \cite{pocock2013predicting}, Twins \cite{almond2005costs} and TCGA \cite{weinstein2013cancer}. Note that these datasets have diverse characteristics in terms of the number of features and the proportion of categorical/continuous features. To the best of our knowledge, none of the medical datasets used contain personally identifiable information nor offensive content. Given that Twins and TCGA are datasets from a single domain, we then explain how we used them to obtain source and target datasets with heterogeneous feature spaces.  

\textbf{MAGGIC.} MAGGIC \cite{pocock2013predicting} consists of datasets with heterogeneous features from 30 domains representing medical studies of patients who have experienced heart failure. Note that this medical dataset is private. Refer to \cite{pocock2013predicting} for details on how the dataset was collected and curated. Each medical study consists of a number of patients ranging from 190 to 13279. From these, we sample a dataset with $<500$ patients to represent our target domain and a dataset with $>500$ patients to represent our source domain. The different datasets in MAGGIC consist of a mixture of categorical and continuous features representing clinical covariates for patients who have experienced heart failure such as age, gender, serum creatinine, diabetes, beta-blocker prescription, lower systolic blood pressure, lower body mass, time since diagnosis, current smoker, chronic obstructive pulmonary disease, etc. The total number of features across the different medical studies in MAGGIC is 216 and the average number of features in a single study is 73. After sampling the source and target dataset from the different medical studies in MAGGIC, we simulate treatment assignments and outcomes as described in Section \ref{sec:evaluation}. Note that different source and target datasets are sampled for each of the different 10 random seeds used for all experimental results.  

\textbf{Twins.} Twins \cite{almond2005costs} represents one of the standard benchmark datasets in the causal inference literature for estimating the effects of binary treatments. The dataset consists of data from 11400 pairs of twin births in the USA recorded between 1989-1991 \cite{almond2005costs}. Refer to \cite{almond2005costs} for details of how the dataset was collected and curated. We use the publicly available version of the dataset from: \url{https://github.com/AliciaCurth/CATENets}. For each twin pair, $39$ relevant covariates were recorded related to the parents, pregnancy and birth such as marital status, mother's age, number of previous births, pregnancy risk factors, quality of care during pregnancy, number of gestation weeks, etc. These represent a mixture of continuous and categorical features. We obtain our full dataset with $N_F = 114000$ examples and $D_F=39$ features by randomly sampling one of the twins to observe. We then sub-sample our source and target datasets as described below and simulate treatments and outcomes as described in Section \ref{sec:evaluation}.

\textbf{TCGA.}  TCGA is a dataset consisting of cancer patients for which gene expression measurements were recorded \cite{weinstein2013cancer}. The dataset is publicly available and has been used by previous methods in causal inference \cite{schwab2019learning, bica2020scigan, kaddour2021causal}. In particular, we use the publicly available version of the TCGA dataset as used by DRNet \cite{schwab2019learning} \url{https://github.com/d909b/drnet}. Refer to \cite{weinstein2013cancer} for details of how the dataset has been collected and curated. The dataset consists of $N_F = 9659$ patients (samples), for which we use the measurements from the $D_F=100$ most variable genes. Note that these are all continuous features. Moreover, we log-normalized the gene expression data and scale each feature in the $[0, 1]$ interval. We then sub-sample our source and target datasets as described below. The treatment assignment and outcomes are simulated as described in Section \ref{sec:evaluation}.

\textbf{Obtaining source and target datasets for Twins and TCGA.}
Unless the size of the target dataset is fixed, from the full datasets, we sample a target dataset of size $N_T \sim \mathcal{U}(100, 500)$ and a source dataset of size $N_R \sim \mathcal{U}(1000, N_F-N_T)$, where $N_T$ is the size of the target dataset, $N_R$ is the size of the source dataset and $N_F$ is the size of the full dataset. For the experiments where we vary the size of the target dataset, we set $N_T \in \{100, 200, 300, 500, 1000, 2000, 4000\}$ and we again sample the size of the source dataset from $N_R \sim \mathcal{U}(1000, N_F-N_T)$. Moreover, to create heterogeneous feature spaces for the source and target domains, let $D_F$ be the number of features in the full dataset. From these, we randomly sample $D_{p_R} \in \mathcal{U}(5, D_{F}/3)$ features that are private for the source dataset, $D_{p_T} \in \mathcal{U}(5, D_{F}/3)$ features that are private for the target dataset and $D_{s} \in \mathcal{U}(5, D_{F}/3)$ features that are shared between the two. Note that different sizes for the source and target datasets and different sets of shared and private features are sampled for each of the different 10 random seeds used for all experimental results.

We use the full source datasets for training the benchmarks making use of the source dataset. Each target dataset undergoes a split into 56/24/20\% for training, validation and testing respectively. The validation dataset is used for early stopping. 

\textbf{Summary statistics for the datasets} We provide in Table \ref{tab:dataset_details} summary statistics of the different characteristics of the full source and target datasets used across our 10 experimental runs. Note that for the experiments where we vary the size of the target dataset, $N_T$ is fixed as described above. To highlight the diversity of the datasets used for evaluating the benchmarks, we highlight in Table \ref{tab:dataset_details} the mean and standard deviations for the size of the source dataset $N_R$, size of the target dataset $N_T$, number of features shared between the source and target datasets $D_s$, number of features private to the source dataset $D_{p_R}$, number of features private to the target dataset $D_{p_T}$, proportion of shared features in the source dataset $D_s / (D_s + D_{p_R})$, proportion of shared features in the target dataset $D_s / (D_s + D_{p_T})$ and the Maximum Mean Discrepancy (MMD) (computed using RBF kernel) between the shared features of the source and target datasets $\text{MMD}(X^{s_R}, X^{s_T})$. Note that the MMD between the shared features of the source and target datasets for MAGGIC is much higher than for TCGA and Twins, as the different datasets in MAGGIC represent medical studies with different patient populations.

\begin{table}[H]
	\centering
	\vspace{-0.4cm}
	\caption{Summary statistics of the datasets sampled for training the benchmarks.}
	\begin{adjustbox}{max width=\textwidth}
	\begin{tabular}{l|ccc}
			 &  TCGA & Twins  & MAGGIC  \\
			\midrule
			$N_R$  & $4219 \pm 2569$ &   $5106 \pm 2988$  &  $2533 \pm 3711$   \\
			$N_T$  & $360 \pm 120$ &   $360 \pm 120$ & $267 \pm 75$  \\
			\midrule
			$D_s$ & $20 \pm 8$ &  $9 \pm 2$ &  $27 \pm 7$    \\
			$D_{p_R}$ & $15 \pm 7$ &  $9 \pm 2$ &   $25 \pm 9$    \\
			$D_{p_T}$ & $21 \pm 9$ &   $8 \pm 3$  & $21 \pm 15$ \\
			$D_s / (D_s + D_{p_R})$ &  $0.57 \pm 0.17$ &  $0.49 \pm 0.10$  &   $0.54 \pm 0.13$   \\
			$D_s / (D_s + D_{p_T})$ & $0.50 \pm 0.14$  & $0.52 \pm 0.08$ & $0.60 \pm 0.19$    \\
			\midrule
			$\text{MMD}(X^{s_R}, X^{s_T})$ &   $0.0009 \pm 0.0005$ &  $0.0011 \pm 0.0008$  &  $0.056 \pm 0.057$  \\
			\bottomrule
	\end{tabular}
	\end{adjustbox}
	\label{tab:dataset_details}
	\vspace{-0.5cm}
\end{table}

\subsection{Implementation details and hyperparameter setting}

As mentioned in Section \ref{sec:evaluation}, we evaluate the following benchmarks (1) training the CATE-learners only on the target dataset, (2) using only the shared features between the source and target datasets and (3) using RadialGAN \cite{yoon2018radialgan} to `translate' the source dataset into the target domain (4) training the HTCE-learners on the full source and target datasets. We describe here the implementation details for RadialGAN \cite{yoon2018radialgan}, for the different CATE-learners used (S-Learner, T-Learner, DR-Learner and TARNet) and the corresponding HTCE-learners (HTCE-S-Learner, HTCE-T-Learner, HTCE-DR-Learner and HTCE-TARNet). 

\noindent\textbf{RadialGAN.} We use the publicly available implementation of RadialGAN \cite{yoon2018radialgan}\footnote{\url{https://github.com/vanderschaarlab/mlforhealthlabpub/tree/main/alg/RadialGAN}} which we augment to the treatment effects setting by considering the treatment separate from the rest of the features. While RadialGAN, was developed to work with any number of source datasets and uses multiple GAN architectures to learn to translate from any one of the source datasets to the target domain, we consider here a simplified case with one source dataset. We use the same hyperparameter setting described in \cite{yoon2018radialgan} and found in the publicly available implementation for the generators, discriminators, encoders and decoders architectures and optimization  (optimizer, batch size, learning rate). 

\noindent\textbf{CATE-learners.} We use the publicly available implementation of the different CATE learners from \cite{curth2021nonparametric}\footnote{\url{https://github.com/AliciaCurth/CATENets}}. This allows us to have a consistent implementation for the S-Learner, T-Learner, DR-Learner and TARNet \cite{shalit2017estimating}. Moreover, we use components similar to those in \cite{curth2021nonparametric, shalit2017estimating} for all networks which ensure that $\hat{\mu}_w(x)$, $\hat{\pi}(x)$ and $\hat{\tau}(x)$ have access to similar number of layers and units in total, thus resulting in equally complex nuisance functions. In particular, for the S-learner, $\hat{\mu}(x, w)$ consists of 5 layers of 200 units each, for the T-learner, each of $\hat{\mu}_0(x), \hat{\mu}_1(x)$ consists of 5 layers with 200 units each. For the DR-Learner, we set each of $\hat{\mu}_0(x), \hat{\mu}_1(x), \hat{\pi}(x), \hat{\tau}(x)$ to have 5 layers with 200 units each. Finally, TARNet consists of 3 representation layers of 200 units each, followed by two outcome heads, each with 2 layers of 100 units each. We use ReLU activation function for the hidden layers and linear/sigmoid activation for the last layer (for continuous/binary outcomes respectively). The different CATE learners are trained using the Adam optimizer \cite{kingma2014adam}, with learning rate set to $0.0001$ and batch size $128$. We perform early stopping on the validation dataset from the target domain.

\noindent\textbf{HTCE-learners.} Each $\phi^{p_R}, \phi^{p_T}, \phi^{s}$ for handling the heterogeneous feature spaces consists of one hidden layer with 100 units each. For the HTCE-S-Learner, we consider a building block of 5 layers with 100 hidden units for each shared and private subspace to share information between $\hat{\mu}^R(x^R, w), \hat{\mu}^T(x^T, w)$. Note that this essentially gives us 200 hidden units for the hypothesis function in each domain, thus ensuring that the nuisance functions for the CATE vs. HTCE-learners in each domain have similar complexity. For the HTCE-T-learner, we use separate feature encoders for each treatment $w\in \{0, 1\}$ ($\phi_w^{p_R}, \phi_w^{p_T}, \phi_w^{s}$ each consisting of one hidden layer with 100 units), and we use two building blocks of 5 layers with 100 units for each shared and private subspace to share information between $(\hat{\mu}^R_0(x^R), \hat{\mu}^T_0(x^T))$ and $(\hat{\mu}^R_1(x^R), \hat{\mu}^T_1(x^T))$. Similarly, for the HTCE-DR-learner, we use the same architecture as for the HTCE-T-learner to obtain the plug-in estimates of the potential outcomes, which consists of separate feature encoders for each treatment $w$, and two building blocks of 5 layers with 100 units for each shared and private subspace to share information between $(\hat{\mu}^R_0(x^R), \hat{\mu}^T_0(x^T))$ and $(\hat{\mu}^R_1(x^R), \hat{\mu}^T_1(x^T))$. Moreover, for the propensity estimation, we use separate feature encoders ($\phi_{\pi}^{p_R}, \phi_{\pi}^{p_T}, \phi_{\pi}^{s}$ each consisting of one hidden layer with 100 units) and a building blocks of 5 layers with 100 units for each shared and private subspace to share information between $(\hat{\pi}^R(x^R), \hat{\pi}^T(x^T))$. Finally, for the pseudo-outcome regressor we also use separate feature encoders ($\phi_{y}^{p_R}, \phi_{y}^{p_T}, \phi_{y}^{s}$ each consisting of one hidden layer with 100 units) and a building blocks of 5 layers with 100 units for each shared and private subspace to share information between $(\hat{\tau}^R(x^R), \hat{\tau}^T(x^T))$. For HTCE-TARNet, we use feature encoders shared between the two treatments ($\phi^{p_R}, \phi^{p_T}, \phi^{s}$ each consisting of one hidden layer with 100 units), followed by 3 representation layers (for each of $\Omega^{R}$, $\Omega^{T}$) and two building blocks of 2 layers with 100 units for each shared and private subspace to share information between $(\hat{\mu}^R_0(x^R), \hat{\mu}^T_0(x^T))$ and $(\hat{\mu}^R_1(x^R), \hat{\mu}^T_1(x^T))$. We use ReLU activation function for the hidden representation layers, SeLU activation for the shared and private subspace layers and linear/sigmoid activation for the last layer (for continuous/binary outcomes respectively). Note that for the building block of sharing information between outcome functions, we build upon the FlexTENet implementation \cite{curth2021inductive}. The different HTCE-learners are trained using the Adam optimizer \cite{kingma2014adam}, with learning rate set to $0.0001$ and batch size $128$. Moreover, we also use the validation dataset from the target domain for early stopping. 

All of the experiments were run on a virtual machine with 6CPUs, an Nvidia K80 Tesla GPU and 56GB of RAM.

\newpage
\section{Additional experimental results}\label{apx:additional_experiments}

This section provides results for the different experimental settings considered in Section \ref{sec:evaluation} for the remaining datasets and for all learners. Note that we observe similar trends for the different experimental settings across all datasets.

\subsection{Benchmarks comparison and source of gain}

Table \ref{tab:benchmars_and_source_of_gain_apx} reports results on the TCGA and MAGGIC datasets for the comparison of the HTCE-learners against the benchmarks and the source of gain analysis. To reiterate, we compare the following benchmarks  (1) training the CATE learners only on the target dataset, (2) using only the shared features between the source and target datasets to train the CATE learners (3) using RadialGAN \cite{yoon2018radialgan} to `translate' the source dataset into the target domain and training the CATE learners on the target dataset augmented in this way and (4) training our HTCE-learners on the full source and target datasets. Moreover, we evaluate the impact of removing the shared and private layers that enable sharing information between the PO functions across domains (HTCE - PO sharing), removing the orthogonal regularization loss for the learnt shared and private feature representations in Equation \ref{eq:loss_orthogonal_rep} (HTCE - $\mathcal{L}_{\text{orth}_{z}}$) and removing orthogonal regularization loss from the PO layers in Equation \ref{eq:loss_orthogonal_po} (HTCE - $\mathcal{L}_{\text{ortho}_{\text{PO}}}$). We note that on the TCGA and MAGGIC datasets as well, the HTCE-learners achieve better performance compared to the baselines and that each component we propose brings performance improvements. 

\begin{table}[h]
	\centering
	\vspace{-0.4cm}
	\caption{Benchmarks comparison and source of gain analysis in terms of PEHE on TCGA and MAGGIC datasets.}
	\begin{adjustbox}{max width=\textwidth}
	\begin{tabular}{l|cccc|cccc}
	        & \multicolumn{4}{c}{TCGA} & \multicolumn{4}{c}{MAGGIC} \\
			Learner &  S-Learner & T-Learner  & DR-Learner & TARNet  &  S-Learner & T-Learner  & DR-Learner & TARNet  \\
			\midrule
			Target & $0.15 \pm 0.02$ &  $0.12 \pm 0.01$ &  $0.10 \pm 0.01$ &  $0.11 \pm 0.01$ & $0.39 \pm 0.04$ &  $0.21 \pm 0.01$ &  $0.19 \pm 0.01$ &  $0.19 \pm 0.01$   \\
			Shared features & $0.30 \pm 0.07$ &  $0.30 \pm 0.07$ &  $0.29 \pm 0.07$ &  $0.30 \pm 0.07$ & $0.32 \pm 0.03$ &  $0.28 \pm 0.03$ &  $0.27 \pm 0.03$ &  $0.27 \pm 0.03$    \\
			RadialGAN & $0.12 \pm 0.01$ &  $0.11 \pm 0.01$ &  $0.09 \pm 0.01$ &  $0.13 \pm 0.03$ & $0.31 \pm 0.01$ &  $0.19 \pm 0.01$ &  $0.19 \pm 0.02$ &  $0.19 \pm 0.01$  \\
			\midrule
			HTCE - PO sharing & $0.15 \pm 0.03$ &  $0.10 \pm 0.01$ &  $0.09 \pm 0.01$ &  $0.10 \pm 0.01$ & $0.29 \pm 0.02$ &  $0.15 \pm 0.01$ &  $0.14 \pm 0.01$ &  $0.17 \pm 0.01$    \\
			HTCE - $\mathcal{L}_{\text{orth}_{z}}$ & $0.10 \pm 0.01$ &  $0.07 \pm 0.01$ &  $0.06 \pm 0.01$ &  $0.07 \pm 0.01$ & $0.25 \pm 0.02$ &  $0.10 \pm 0.01$ &  $0.09 \pm 0.01$ &  $0.12 \pm 0.01$   \\
			HTCE - $\mathcal{L}_{\text{ortho}_{\text{PO}}}$ & $0.11 \pm 0.01$ &  $0.08 \pm 0.01$ &  $0.07 \pm 0.01$ &  $0.05 \pm 0.01$ & $0.26 \pm 0.02$ &  $0.10 \pm 0.01$ &  $0.10 \pm 0.01$ &  $0.12 \pm 0.01$   \\
			\midrule
			HTCE (ours) & $\textbf{0.07} \pm 0.01$ &  $\textbf{0.06} \pm 0.01$ &  $\textbf{0.04} \pm 0.01$ &  $\textbf{0.06} \pm 0.01$ &  $\textbf{0.24} \pm 0.02$ &  $\textbf{0.08} \pm 0.01$ &  $\textbf{0.08} \pm 0.01$ &  $\textbf{0.10} \pm 0.01$  \\
			\bottomrule
	\end{tabular}
	\end{adjustbox}
	\label{tab:benchmars_and_source_of_gain_apx}
\end{table}

When looking at the performance of the benchmarks, using only the shared features between the two datasets generally has poor performance. This is due to the fact that this introduces hidden confounders in the CATE estimation since all of the patient features in both domains are used to obtain the potential outcomes and the treatment assignment in our data simulation (see Section \ref{sec:evaluation}). Moreover, note that RadialGAN only slightly improves performance over only training the CATE learners on the target dataset. We hypothesise that this may also be due to the small size of the target dataset which hinders the generator networks in RadialGAN to learn how to map examples from the source domain to the target one.

\subsection{Varying the information sharing between domains} 

\begin{figure}[h]
    \centering
    \subfloat{\includegraphics[height=2.5cm, clip]{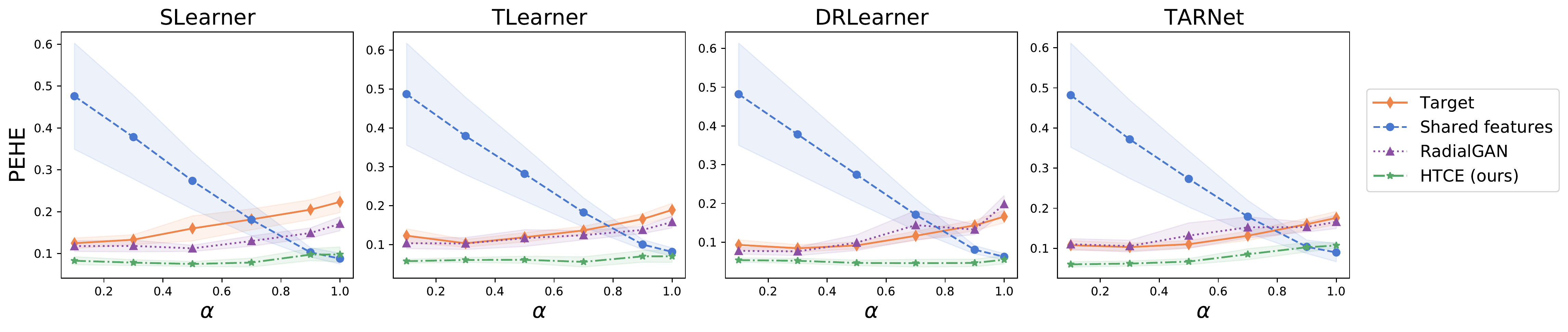}} 
    
    \subfloat{\includegraphics[height=2.5cm, clip]{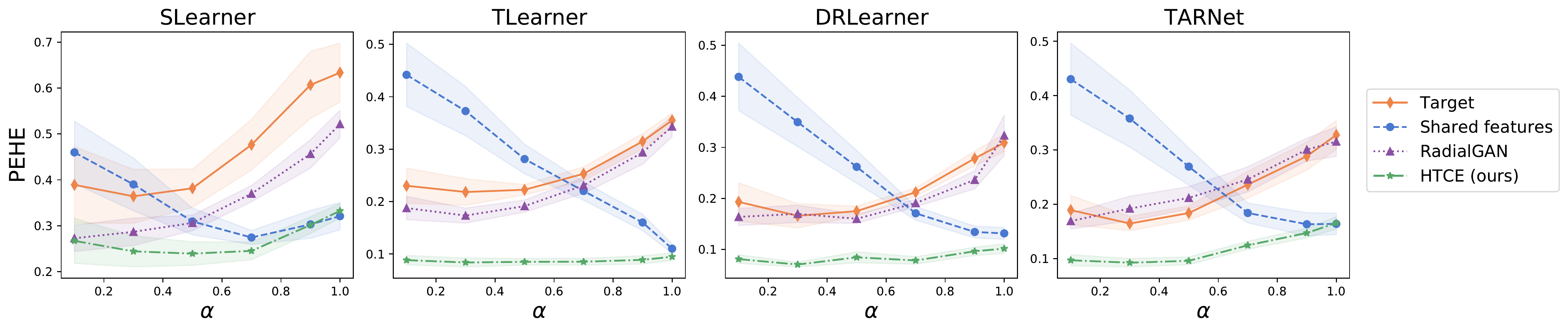}}
    \caption{Performance comparison on TCGA (top) and MAGGIC (bottom) when varying the information sharing between PO functions across domains through $\alpha$.}
    \label{fig:performance_alpha_apx}
\end{figure}

Figure \ref{fig:performance_alpha_apx} reports the results for varying the parameter $\alpha$, which controls the amount of information shared between the PO in the source and target domains, on the TCGA (top) and MAGGIC (bottom) datasets. We again notice that our HTCE-learners, which use a flexible approach for information sharing between PO functions across domains achieve good performance both when the PO functions are significantly different between the source and target dataset ($\alpha=0.1$) and also when they have the same functional form and only depend on the shared features ($\alpha=1.0$). In addition, for high values of $\alpha$ and especially for $\alpha=1.0$ we again notice the degradation in performance due to the fact that the learners need to perform the implicit task of feature selection.

\subsection{Varying the target dataset size.}

\begin{figure}[h]
    \centering
    \includegraphics[height=2.5cm, clip]{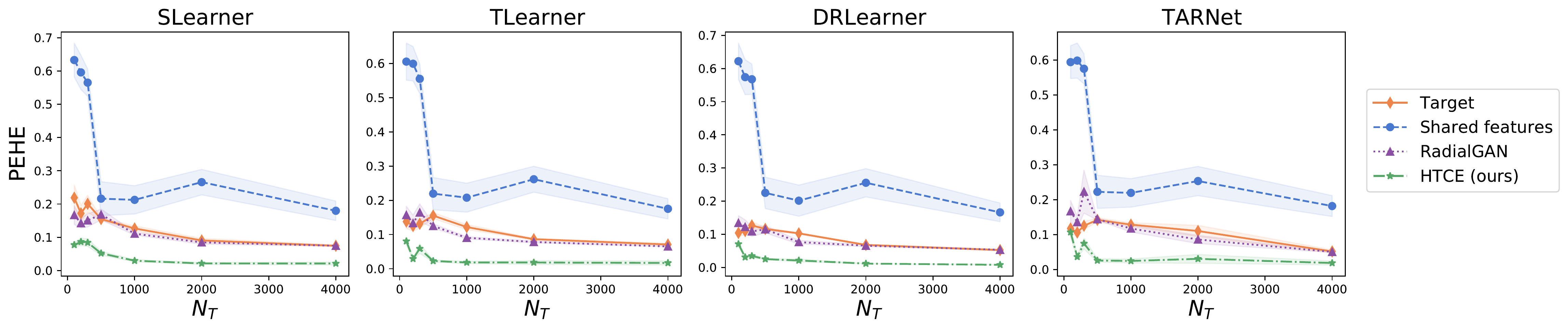} 
    \caption{Performance comparison on TCGA when varying the the size of the target dataset ($N_T$).}
    \label{fig:performance_alpha_size_target_apx}
\end{figure}

Figure \ref{fig:performance_alpha_size_target_apx} reports the results on when varying the size of the target dataset $N_T$ on the TCGA dataset. Note that for the MAGGIC dataset, the size of the target domain is not simulated and it represents the size of the various studies selected as target domains. Thus, we only report here results on  TCGA where we directly simulate the size of the target dataset. We notice that the benefits of doing transfer learning degrade as we increase the size of the target dataset.

\subsection{Effect of selection bias} 

\begin{figure}[H]
    \centering
    \subfloat[SLearner]{\includegraphics[height=2.3cm, clip]{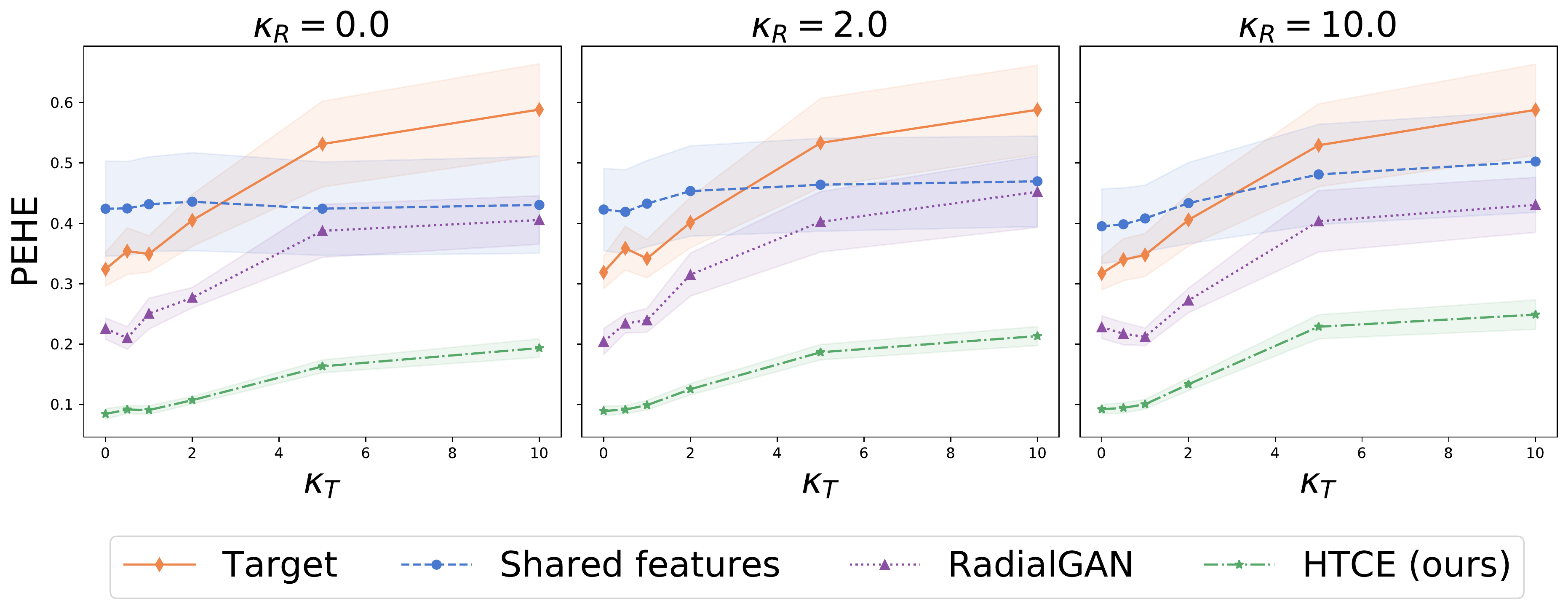}} 
    \subfloat[TLearner]{\includegraphics[height=2.3cm, clip]{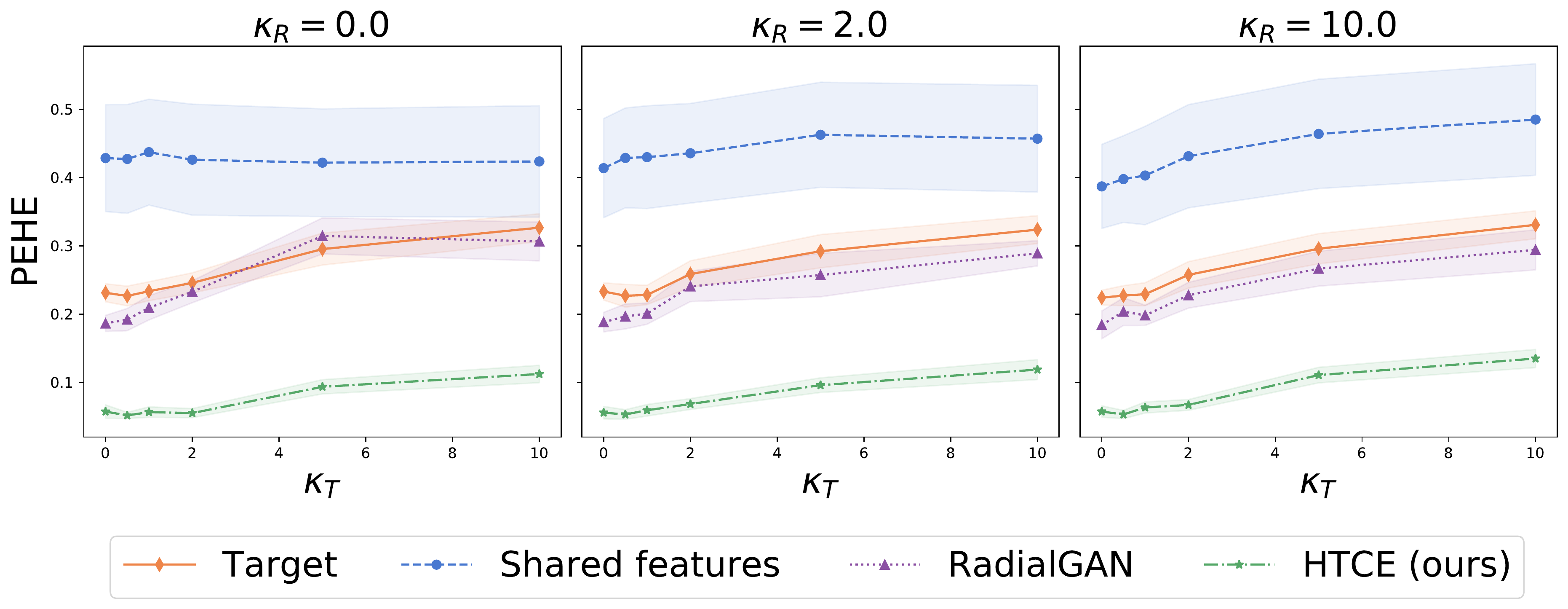}}
    
    \subfloat[DRLearner]{\includegraphics[height=2.3cm, clip]{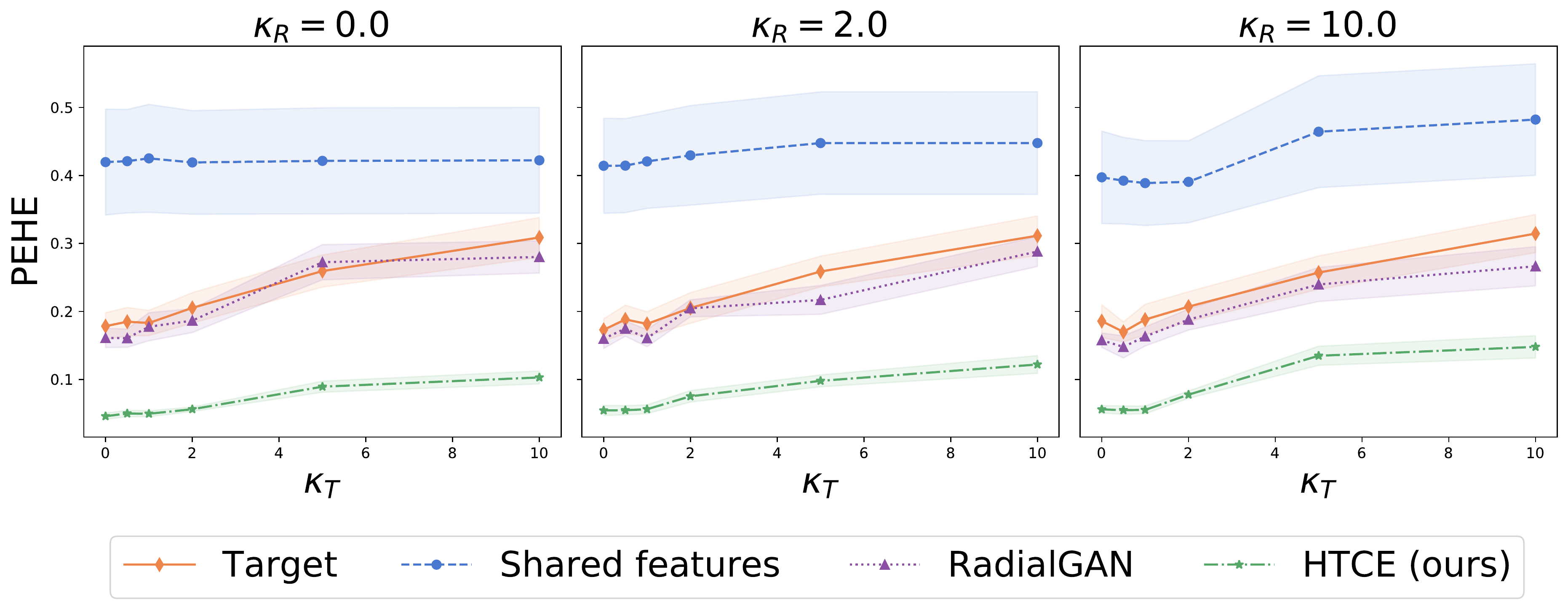}} 
    \subfloat[TARNet]{\includegraphics[height=2.3cm, clip]{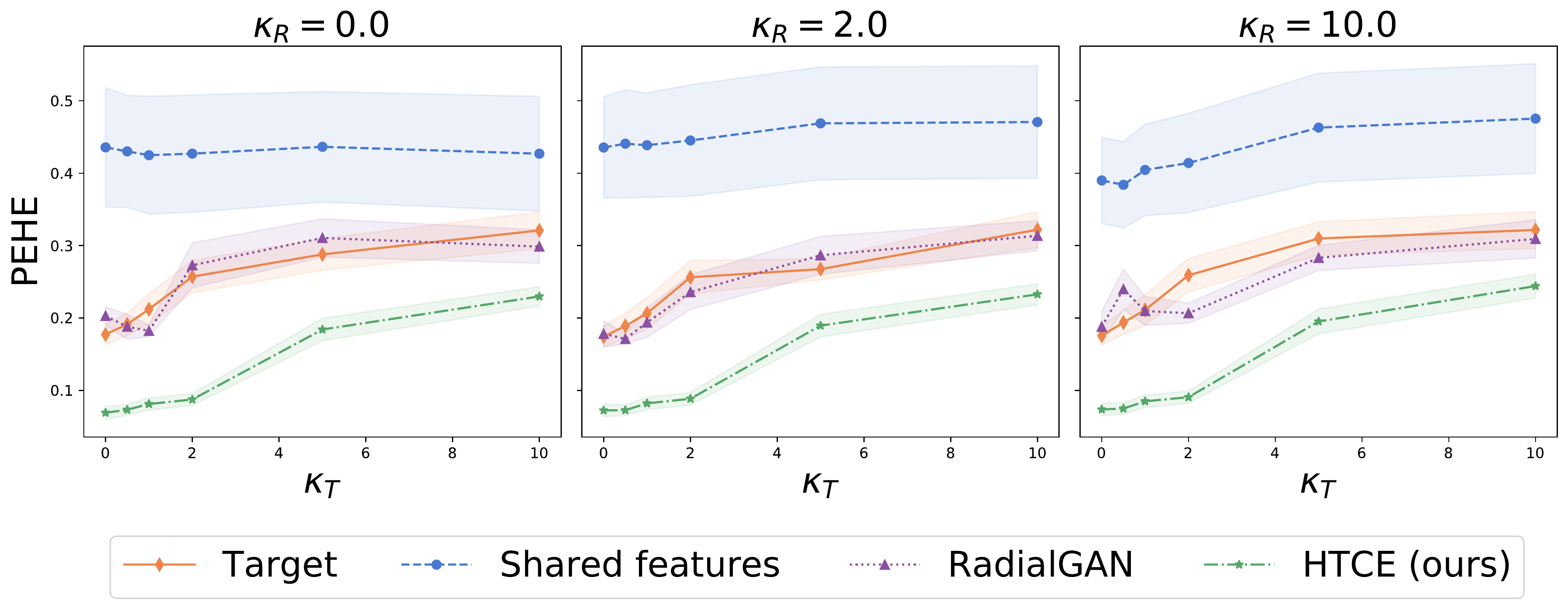}}
    \caption{Performance comparison on Twins for all learners when varying the selection bias $\kappa_R$ and $\kappa_T$ in the source and target datasets.}
    \label{fig:performance_selection_bias_twins}
    \vspace{-5mm}
\end{figure}

\begin{figure}[H]
    \centering
    \subfloat[SLearner]{\includegraphics[height=2.3cm, clip]{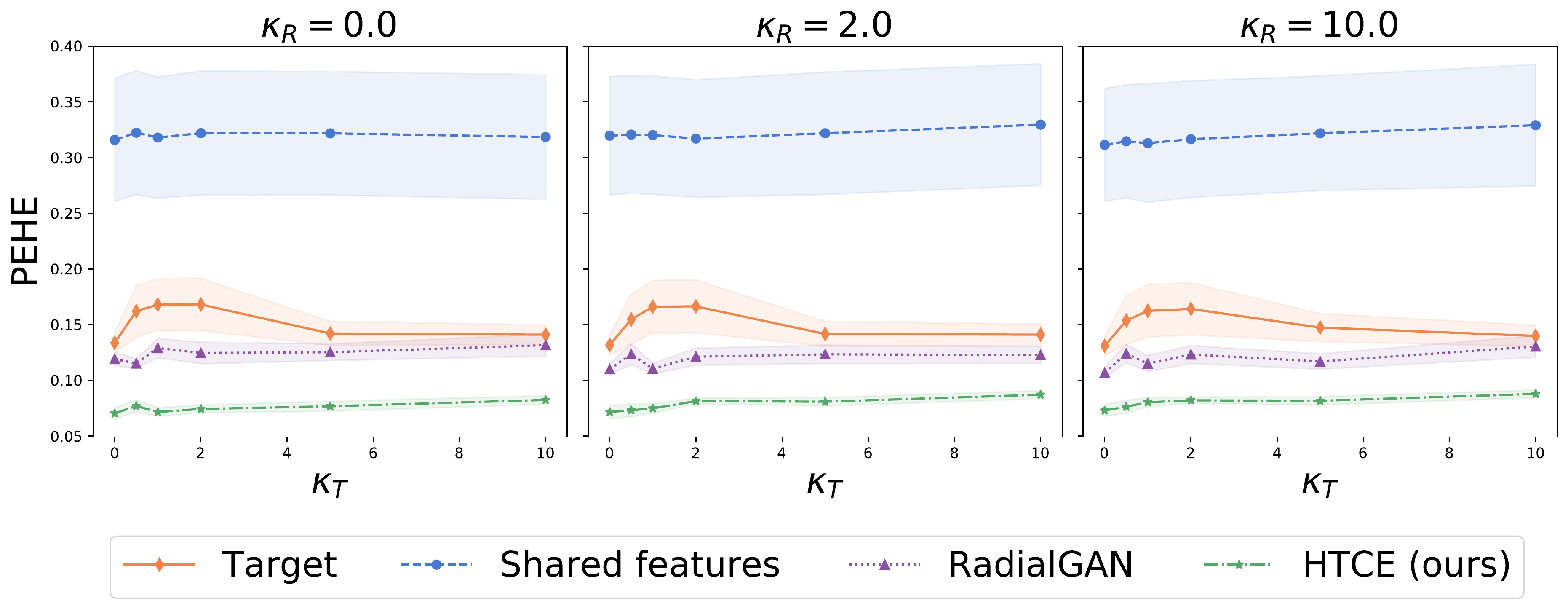}} 
    \subfloat[TLearner]{\includegraphics[height=2.3cm, clip]{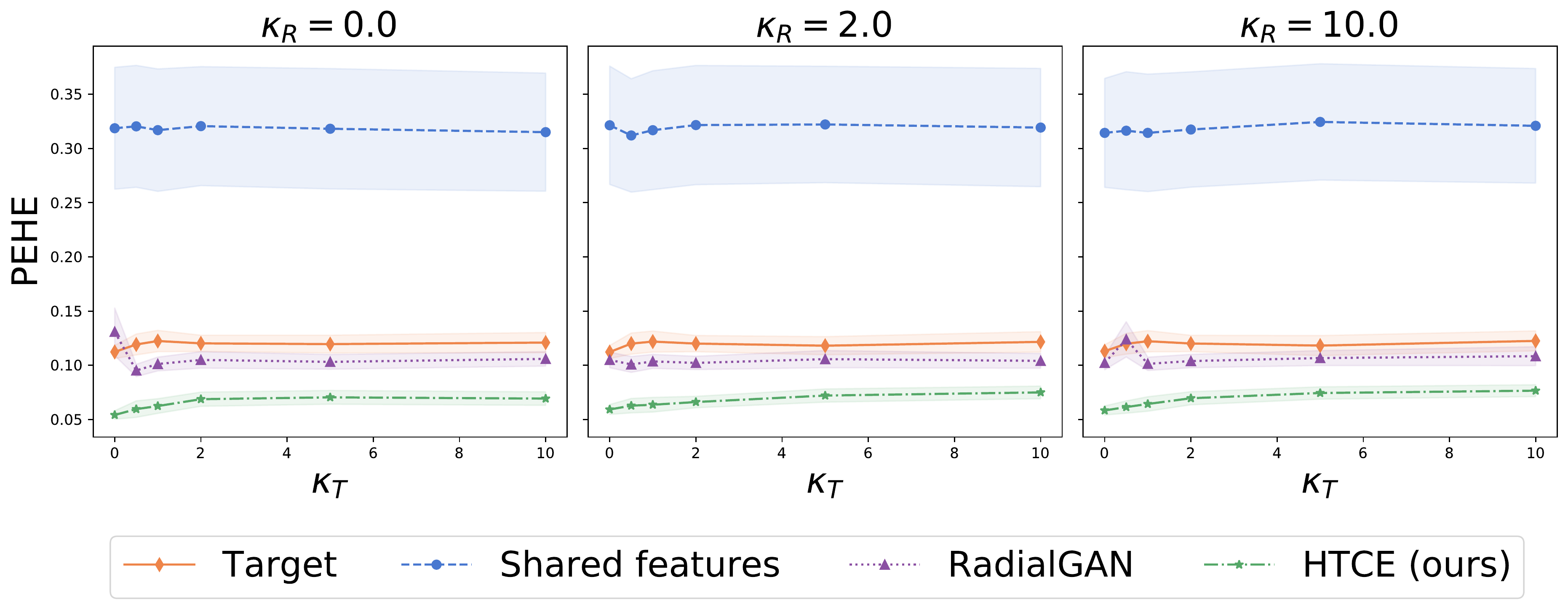}}
    
    \subfloat[DRLearner]{\includegraphics[height=2.3cm, clip]{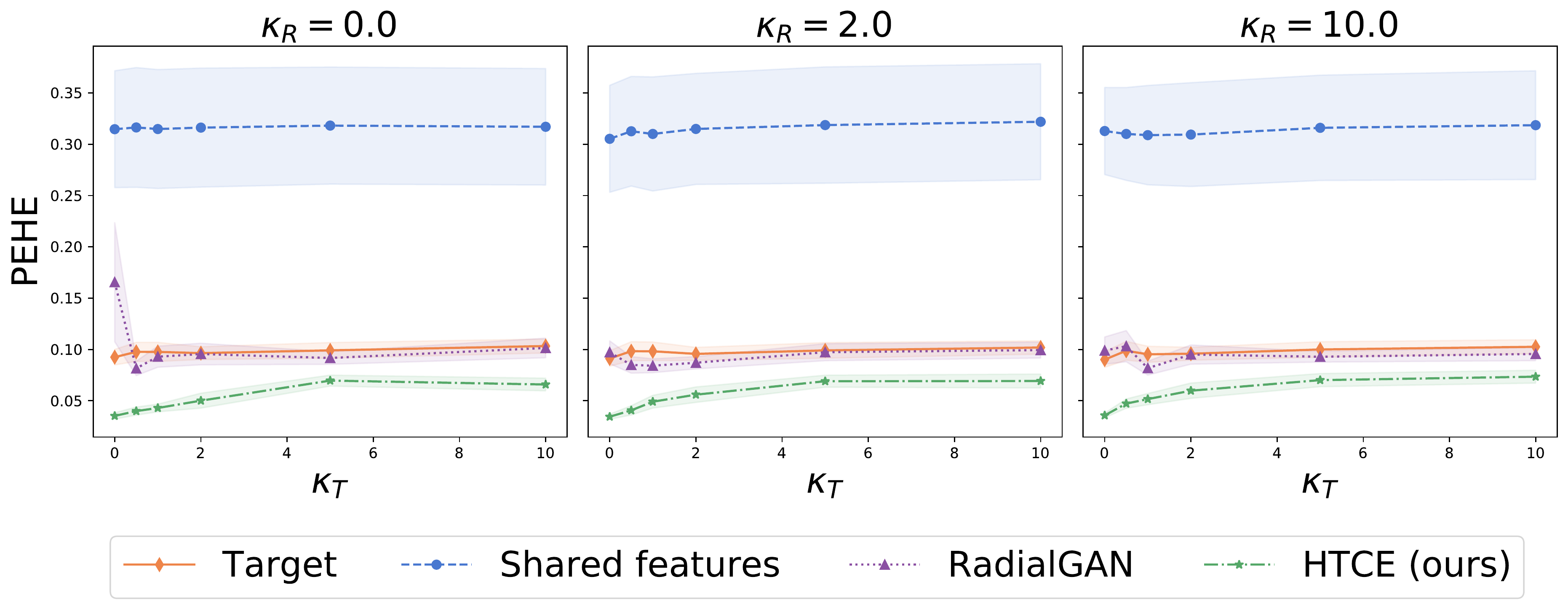}} 
    \subfloat[TARNet]{\includegraphics[height=2.3cm, clip]{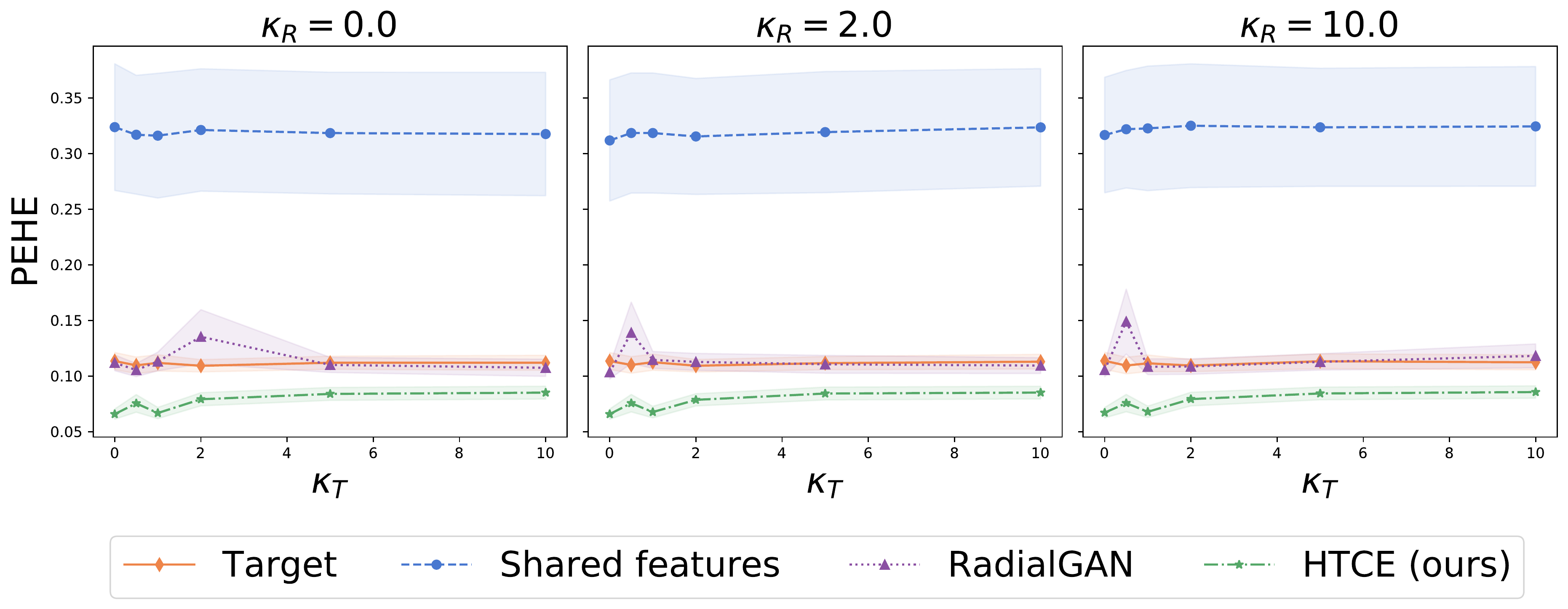}}
    \caption{Performance comparison on TCGA for all learners when varying the selection bias $\kappa_R$ and $\kappa_T$ in the source and target datasets.}
    \label{fig:performance_selection_bias_tcga}
    \vspace{-5mm}
\end{figure}

\begin{figure}[H]
    \centering
    \subfloat[SLearner]{\includegraphics[height=2.3cm, clip]{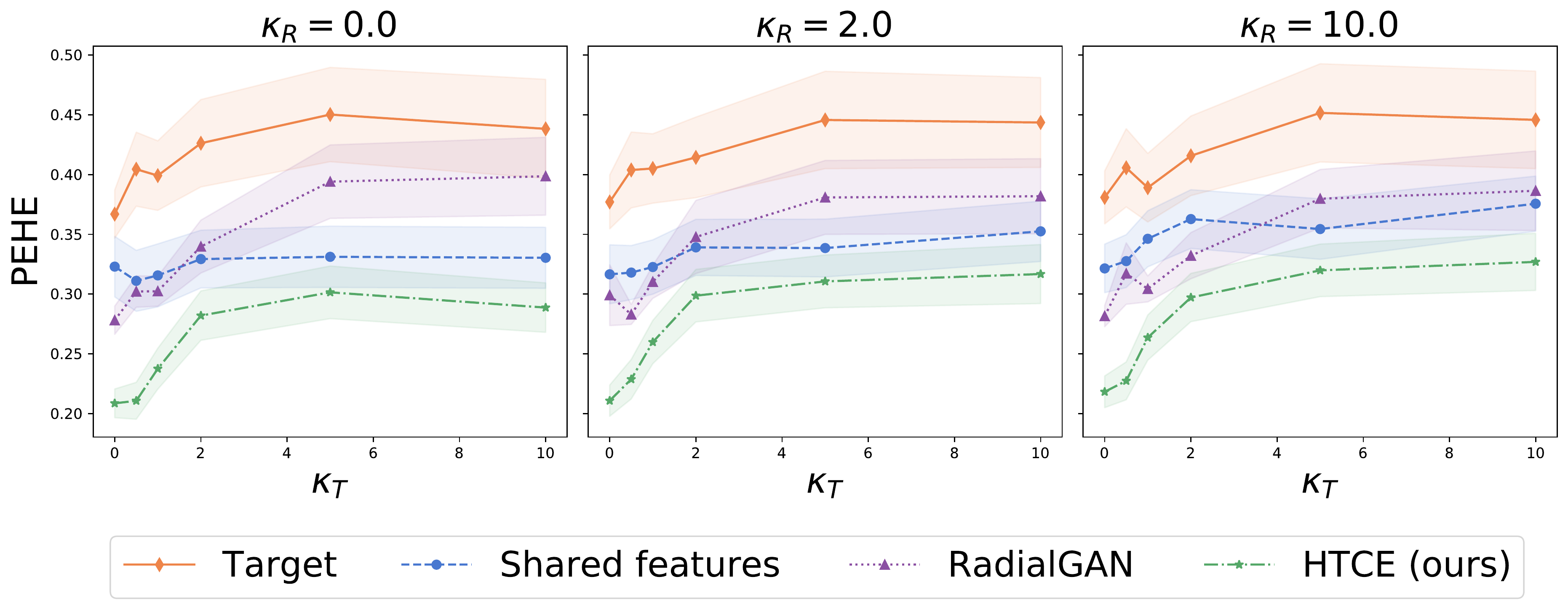}} 
    \subfloat[TLearner]{\includegraphics[height=2.3cm, clip]{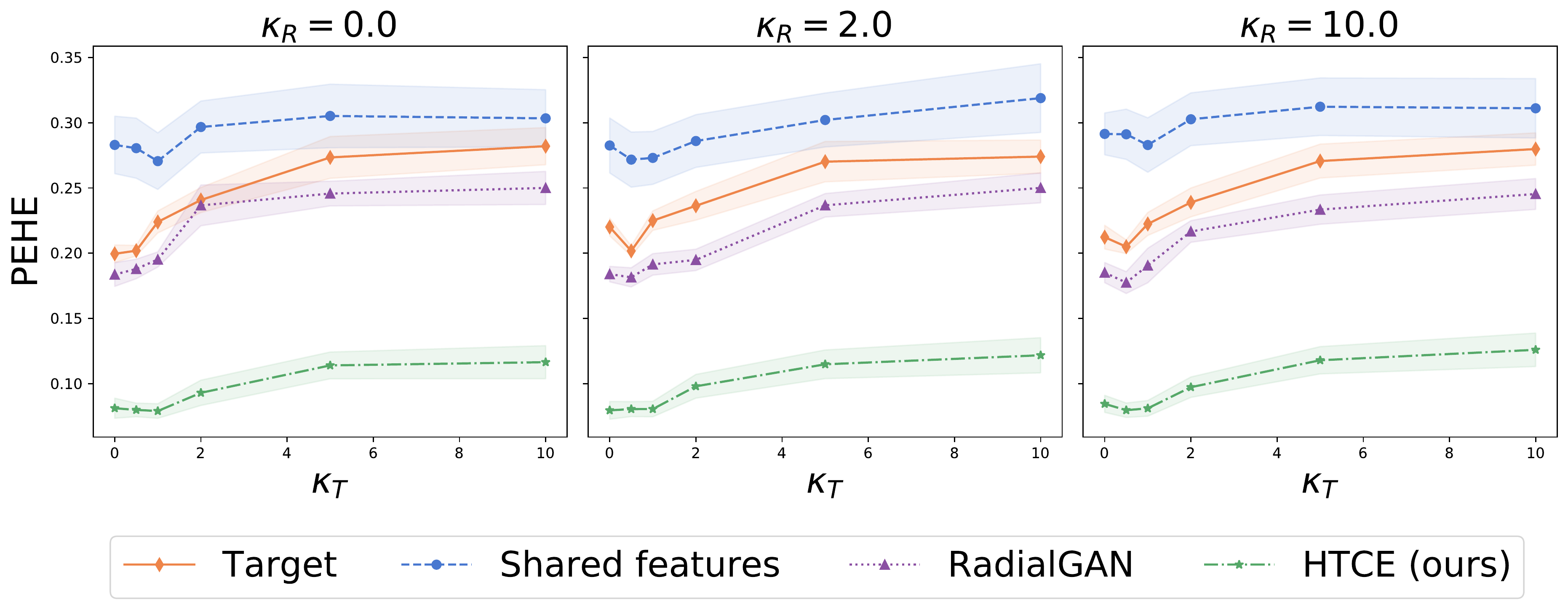}}
    
    \subfloat[DRLearner]{\includegraphics[height=2.3cm, clip]{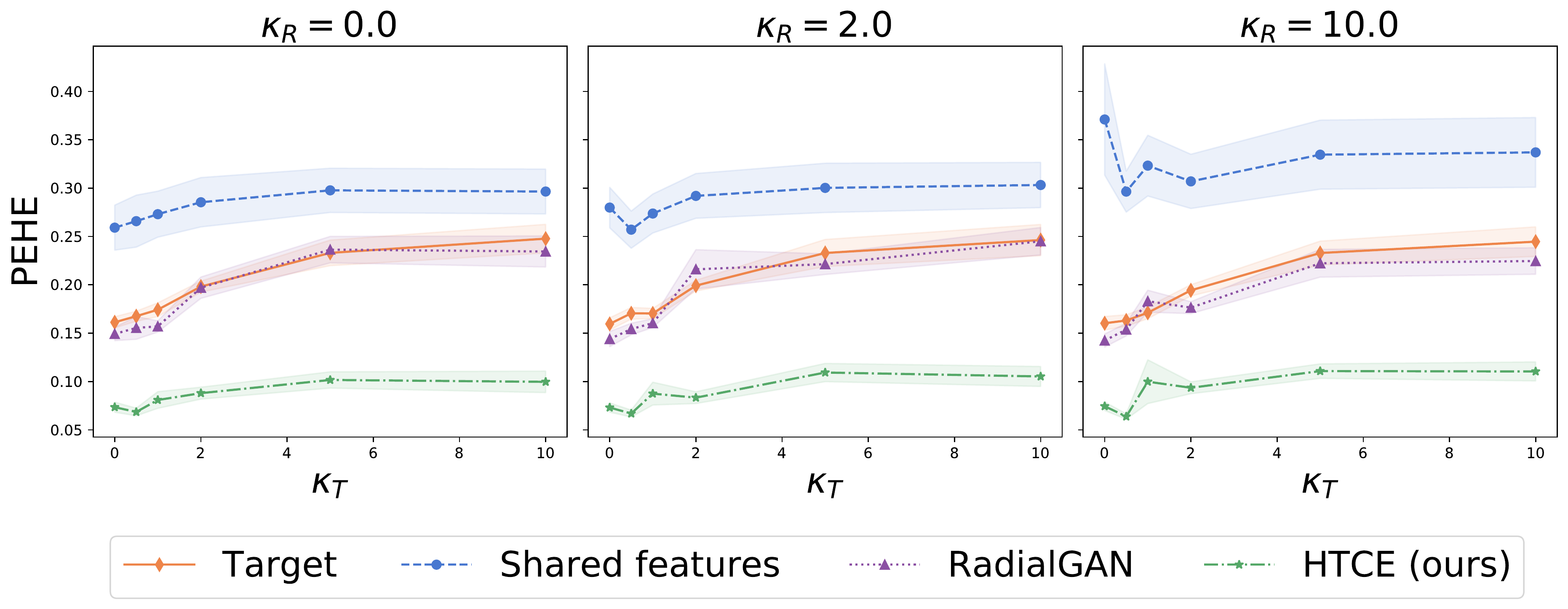}} 
    \subfloat[TARNet]{\includegraphics[height=2.3cm, clip]{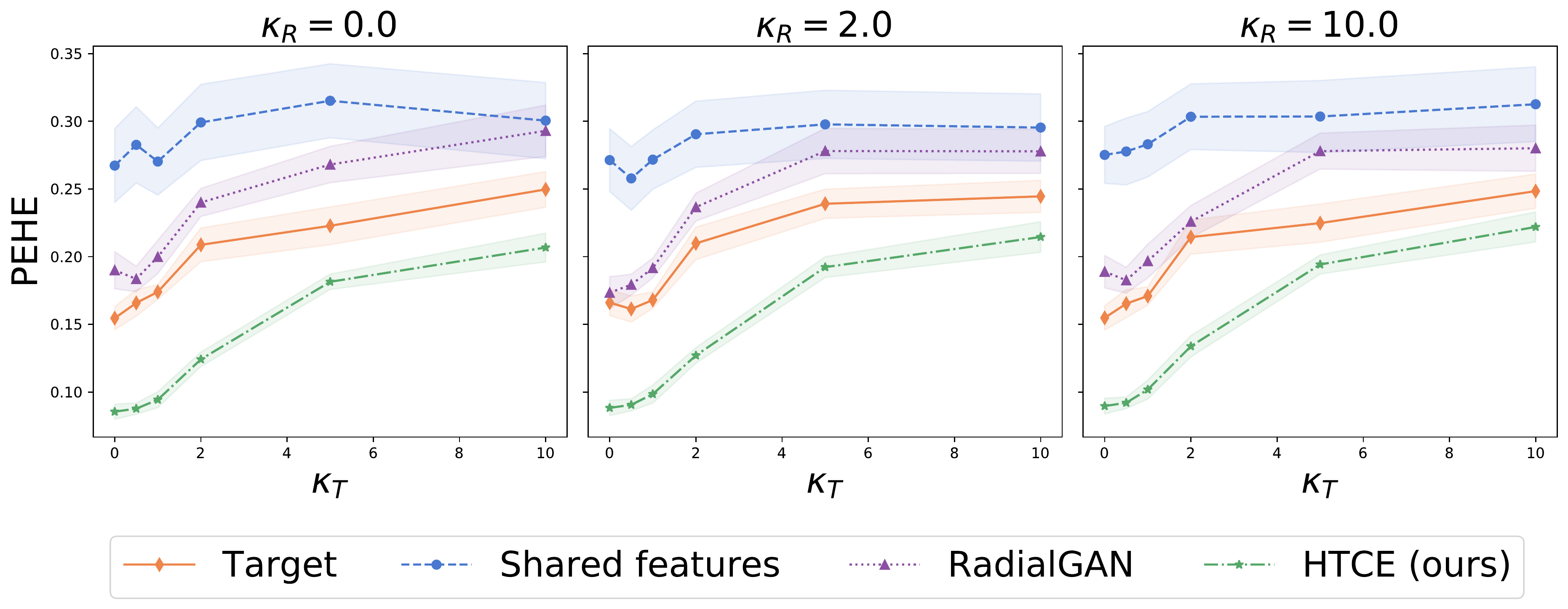}}
    \caption{Performance comparison on MAGGIC for all learners when varying the selection bias $\kappa_R$ and $\kappa_T$ in the source and target datasets.}
    \label{fig:performance_selection_bias_maggic}
\end{figure}

Finally, we report results when varying the selection bias in the source and target domains for all learners on all three datasets: Twins (Figure \ref{fig:performance_selection_bias_twins}), TCGA (Figure \ref{fig:performance_selection_bias_tcga}) and MAGGIC (Figure \ref{fig:performance_selection_bias_maggic}). We consider three setting for the selection bias in the source dataset $\kappa_R = 0.0$ for random treatment assignment, $\kappa_R = 2.0$ for moderate and $\kappa_R = 10.0$ for strong selection bias. For each setting of $\kappa_R$, we vary the selection bias in the target dataset $\kappa_T$ from $\kappa_T = 0.0$ to $\kappa_T = 10.0$. 
Our HTCE-learners have consistent performance when increasing the selection bias in both datasets and when there are significant discrepancies in the treatment assignment mechanism between the source and target datasets. While the impact of increasing the selection bias differs among the learners (e.g. TARNet vs. DRLearner), note that this is due to the intrinsic characteristics of each learner on a single domain, as the DRLearner uses propensity weighting to adjust for the selection bias, while TARNet does not. 

\newpage

\newpage

\section{Additional discussion about extensions and limitations}\label{apx:additional_extensions}

In this section, we describe how our method could be extended to additional scenarios than the ones considered in the paper and further discuss limitations.

 \textbf{Multiple source datasets.} While in this paper we mainly consider the standard and most common transfer learning setting \cite{pan_and_yang_survey} of leveraging a source dataset to improve the estimation of outcomes on a target dataset of interest, our proposed approach could be easily extended to having multiple source domains, and in fact, would scale linearly to incorporating additional domains. 

More precisely, consider access to $M$ source domains, $\{\mathcal{D}^{R_m}\}_{m=1}^{M}$, where each source domain consists of $\mathcal{D}^{R_m} = \{(X^{R_m}_i, W_i, Y_i)\}_{i=1}^{N_{R_m}}$ and a target domain as described in Section \ref{sec:problem_formalism}. The objective would still be the one from Section \ref{sec:problem_formalism} of improving the estimation of CATE for the target domain, but in this case, we need to leverage data from all $M$ source domains. Our proposed building blocks could be extended as follows to handle the additional source domains. The building block for handling heterogeneous feature spaces between source and target domains could be extended by having encoders $\phi^{R_m}$ for the private features $X^{p_{R_m}}$ of each source domain (part of $X^{R_m}$). Moreover, we can build an approach that shares information between the potential outcomes of each source domain with the target domain by having $L$ layers shared between all source domains and the target domains and $L$ private layers for each source domain. For source domain $m$, the input to the layer $(l+1)$-th can be obtained using $\tilde{h}_{w, l+1}^{p_{R_m}} = [ h_{w, l}^{s} || h_{w, l}^{p_{R_m}}]$ where $h_{w, l}^{p_{R_m}}$ is the output of the $l$-th private layer for source domain $m$ and $h_{w, l}^{s}$ is the output of the $l$-th shared layer across all domains. 

While from a model development perspective this extension can be easily done, one also needs to consider whether the multiple source domains satisfy the underlying implicit assumptions for which such an architecture would be appropriate. In particular, it would be important to consider whether the PO functions across all source domains share information with the PO functions in the target domain, as using source domains that are significantly different from the target domain could harm performance.

\textbf{Streaming datasets.} Another important setting to consider is the one of having streaming datasets. In this scenario, one option could be the case where we already have a source domain and a target domain and we have incoming data streaming from the target domain. One such example in healthcare would be the case where in a single hospital we start collecting more or different clinical covariates for the patients which we now want to use for CATE estimation. The source domain would be the patients with only the original set of features, while the target domain would be the patients with the new set of features. However, as we start collecting these additional features, the initial target dataset will be small but it can start increasing with time as we observe more patients. In this setting, we can train an HTCE-learner with the initial data available from the source and target domains. However, instead of retraining the full HTCE-learner as we obtain more patients from the target domain, one option would be to fine-tune the weights using the incoming examples. While this is outside the scope of our paper, we believe that it would be important to investigate appropriate ways for performing such fine-tuning.

Another option could be the case of having full source datasets streaming, while the target dataset remains fixed. This would happen in the setting where for instance we gradually get access to data from multiple locations and we want to use these datasets as source datasets. In this scenario, one possibility would be to use the approach described above for having multiple source datasets and retraining a new model that incorporates all of the available source datasets as we get access to them. Another possibility would be to, instead of retraining a full model as we go from $M$ to $M+1$ source domains, we can add the needed private encoder and layers for the $(M+1)$-th domain and train only the new parameters and fine-tune the shared ones with the data from the new domain. While this is again outside our scope, it can provide interesting avenues for future work.

\textbf{Unknown domains.} In this paper, we assume that the source and target domains are known. However, given that we handle the setting of transfer learning for heterogeneous feature spaces, if the domains are unknown, one way to split available data into different domains, in this case, would be by using the same features to denote a single domain. For instance, if different patients in the dataset have recorded different sets of features, then the patients can be grouped according to having the same information collected for them and these can denote the different domains. Then, one needs to decide which represents the target domain, while keeping in mind that doing transfer learning is most beneficial when the target dataset is small (as highlighted by our experimental results in Section \ref{sec:results_and_discussion} and Figure \ref{fig:performance_alpha_size_target} (bottom)).

\textbf{Personalized feature spaces.} In the case where different features are available for different patients and the source and target domains are unknown, then the patients can be grouped according to their feature spaces as described in the paragraph above. On the other hand, if the source and target datasets are pre-defined and the patients within each dataset have different features available for them, one possible option would be to consider the super-set of their features as the different feature spaces and consider the features that are not available for each patient as missing. However, as our HTCE-learners has not been designed for this particular setting of having missing data, one would also need to investigate if additional assumptions (in addition to ensuring that the no hidden confounders and overlap assumptions are still satisfied) are needed to be able to obtain valid estimates of causal effects.

\end{document}